\documentclass[twoside]{article}

\usepackage[accepted]{aistats2025}
\usepackage[round]{natbib}

\usepackage{enumitem}
\usepackage{multirow}
\usepackage{multicol}
\usepackage{float}
\usepackage{booktabs}
\usepackage{packages}
\usepackage{dblfloatfix}
\newcommand{\ccint}[1]{\left[#1\right]}

\definecolor{hanblue}{rgb}{0.27, 0.42, 0.81}
\definecolor{deepred}{HTML}{900C3F}
\definecolor{deepgreen}{HTML}{2F6960}

\theoremstyle{plain}

\theoremstyle{definition}

\theoremstyle{remark}

\def\pdata{p_{\textup{data}}}

\def\jac{\mathbf{J}}

\def\bfX{\mathbf{X}}

\def\bfB{\mathbf{B}}

\def\mcx{\mathcal{X}}

\def\mcn{\mathcal{N}}

\def\rmd{\mathrm{d}}

\def\Ibb{\mathbb{I}}

\newcommand{\defeq}{\vcentcolon=}

\def\expect{\mathbb{E}}
\def\rset{\mathbb{R}}

\newtcolorbox{mybox}[3][]
{
  colframe = #2!25,
  colback  = #2!10,
  coltitle = #2!20!black,  
  title    = {#3},
  #1,
}

\NewDocumentCommand{\codeword}{v}{%
\texttt{\textcolor{blue}{#1}}%
}

\begin{document}

%
\runningtitle{Distilled Energy Diffusion Models}

%
\runningauthor{Thornton, Béthune, Zhang, Bradley, Nakkiran, Zhai}



\twocolumn[

\aistatstitle{Composition and Control with Distilled Energy Diffusion Models and Sequential Monte Carlo}
\vspace{-0.5cm}
\aistatsauthor{James Thornton  \\ Apple
\And Louis Béthune \\ Apple
\And Ruixiang Zhang \\ Apple}
\aistatsauthor{Arwen Bradley \\ Apple
\And Preetum Nakkiran \\ Apple
\And Shuangfei Zhai \\ Apple}
\vspace{0.3cm}
]

\begin{abstract}
Diffusion models may be formulated as a time-indexed sequence of energy-based models, where the score corresponds to the negative gradient of an energy function. As opposed to learning the score directly, an energy parameterization is attractive as the energy itself can be used to control generation via Monte Carlo samplers. Architectural constraints and training instability in energy parameterized models have so far yielded inferior performance compared to directly approximating the score or denoiser. We address these deficiencies by introducing a novel training regime for the energy function through distillation of pre-trained diffusion models, resembling a Helmholtz decomposition of the score vector field. We further showcase the synergies between energy and score by casting the diffusion sampling procedure as a Feynman Kac model where sampling is controlled using potentials from the learnt energy functions. The Feynman Kac model formalism enables composition and low temperature sampling through sequential Monte Carlo.
\end{abstract}

\section{Introduction}

Diffusion models \citep{sohldickstein2015deep, song2020generative, song2021scorebased} have come to dominate the generative modelling landscape, exhibiting state of the art performance \citep{dhariwal2021diffusion} across domains and modalities \citep{de2022riemannian}, including self-supervised learning \citep{chen2024deconstructing}, natural science applications \citep{arts2023two} and for optimal transport \citep{debortoli2021diffusion}.

Despite the success of diffusion models, there are still a number of challenges. Firstly, diffusion models are known to have slow and expensive training \citep{jeha2024variance, zhang2023improving}. Secondly, effective conditioning of diffusion models remains an open challenge \citep{wu2024practical, zhao2024conditional}. Re-using, fine-tuning and adapting pretrained models has become an active area of research; both to overcome lengthy pretraining and to introduce additional conditioning not considered during training \citep{ye2024tfg, du2021improved}. In addition, many heuristic guidance weighting methods and prompt engineering techniques have been proposed to control generation. Such approaches are poorly understood \citep{bradley2024classifier}, often require ad-hoc weighting and trial-and-error sampling to reach desired samples. 

A diffusion model may be viewed as a sequence of time-indexed energy-based models, where the gradient of the energy is learnt rather than the energy itself \citep{song2021trainebm, salimans2021should}. It was shown in \cite{du2023reduce} how the energy interpretation of diffusion models may be used to better condition and compose pretrained diffusion models in order to generate novel distributions, such as composed distributions, rather than the default reverse process. \cite{du2023reduce}  achieves this through Markov Chain Monte Carlo (MCMC) and annealed Langevin dynamics. Whilst a promising direction, energy-parameterised diffusions are inherently cumbersome to train and annealed MCMC sampling requires an excessive number of network evaluations.

 Energy-parameterized diffusion models require computing multiple gradients through the energy function during training; first with respect to (w.r.t) the input state to recover a score, then w.r.t parameters for training. This exacerbates the already lengthy training entailed by denoising score-matching. 

\textbf{Contributions}
The purpose of this work is two-fold. First to address training instability of energy-parameterized diffusion models, and secondly to introduce a new class of diffusion model samplers using the energy function for controllable generation within a Feynman Kac - Sequential Monte Carlo framework.

We summarize our key contributions as follows:
\begin{itemize}
\item We introduce a novel training procedure and parameterization to efficiently distill pretrained diffusion models into energy based models, whilst avoiding the high-variance loss of denoising score-matching. Our method can be interpreted as a \textit{conservative projection} of a pretrained score.
\item We showcase the merits of our approach in terms of generative performance, consistently achieving superior FID to prior energy-parameterised models for the datasets considered.
\item We describe a general framework of how diffusion models may be used as the underlying Markov process of a Feynman Kac model (FKM); we detail how prior SMC based diffusions may be viewed as particular cases.
\item Finally, we demonstrate how the energy function may be used to construct modality agnostic potentials within FKMs. This enables temperature controlled sampling, as well as composition of diffusion models via SMC. 
\end{itemize}

\section{Background} \label{sec:background}

\subsection{Diffusion Models}
Consider data distribution $\pdata$ on support $\mcx$, and let the stochastic process $(\bfX_t)_{t=0}^T$ be given by the following dynamics; known as the forward process:
    \begin{equation}
          \label{eq:forward}
            \rmd \bfX_t = f(t)\bfX_t \rmd t + g(t) \rmd \bfB_t  ,\quad \bfX_0 \sim p_0\defeq\pdata,
        \end{equation}
        with Brownian motion, $(\bfB_t)_{t \in \ccint{0,T}}$, drift
        $f: \rset \to \rset$ and scale $g: \rset \to \rset$ applied coordinate wise and let $p_t$ denotes the marginal density of $\bfX_t$.

    Diffusion models \citep{song2020denoising, song2021scorebased, ho2020denoising} generate new samples by simulating from the stochastic process \eqref{eq:time_reversed} with density denoted $q^\lambda_{0:T}$, initialization $\tilde{\bfX}_0 \sim p_T$ and $\tilde{\bfB}_t$ denotes another Brownian motion, independent to the forward process. 
    \begin{equation}\small 
      \label{eq:time_reversed}
      \rmd \tilde{\bfX}_t = \left[-f(t) + g^2(t) \tfrac{1+\lambda^2}{2}\nabla \log p_{T-t}(\tilde{\bfX}_t) \right] \rmd t + \lambda g(t)
      \rmd \tilde{\bfB}_t.
    \end{equation}
    Parameter $\lambda>0$ controls the degree of stochasticity
    within $q^\lambda_{0:T}$ \citep{diffnormflow, zhang2022fast}. In particular, for $\lambda=1$, $p_{0:T}=q^1_{0:T}$, hence $q^1_{0:T}$ corresponds to the time-reversal of \eqref{eq:forward} \citep{haussmann1986time,anderson1965iterative}. Setting $\lambda=0$, results in the probability flow ODE \citep{song2021scorebased}. The marginal distributions of \eqref{eq:time_reversed} match those of \eqref{eq:forward}, i.e. $q^\lambda_t = p_t$, for all $t$ and all $\lambda\geq 0$, hence \eqref{eq:time_reversed} generates the data distribution $q^\lambda_0=p_0=\pdata$.

    \textbf{Training.} The score term, $\nabla \log p_{t}(\bfX_t)$, is generally intractable but can be expressed as the solution to a regression problem on the conditional score, then approximated by training a parameterized function $s^*_\theta$. This is known as denoising score-matching (DSM) \citep{song2019generative, vincent2011connection}:
    \begin{align*} 
         s^*_\theta =\arg\min_{s_\theta} \mathbb{E}_{p_{0,t}}[\|s_\theta(\bfX_{t},t) - \nabla_{x_{t}} \log p_{t|0}(\bfX_{t}|\bfX_{0})\|^2].
    \end{align*}
     Given the drift function in \eqref{eq:forward} is typically chosen to be linear in state $\bfX_t$ and applied coordinate-wise, then the forward process may be sampled in closed form using $\bfX_t|x_0 = \alpha_t x_0 + \sigma_t\epsilon$, $\epsilon\sim \mcn(\mathbf{0}, \Ibb)$, for some time-indexed coefficients $\alpha_t, \sigma_t \in \mathbb{R}^+$, \citep{sarkka2019applied, song2021scorebased}. The conditional score $\nabla_{x_{t}} \log p_{t|0}(\bfX_{t}|\bfX_{0})$ is therefore tractable. Alternatively, based on Tweedie's formula \citep{efron_tweedie, Robbins1956AnEB}: $
    \nabla \log p_t(x_t) = \sigma_t^{-2}(\alpha_t \mathbb{E}_{X_0|x_t}[\bfX_0 |x_t] - x_t)$; one may approximate the expected denoiser $\mathbb{E}_{\bfX_0|x_t}[\bfX_0 |x_t]$ via regression as in \eqref{eq:denoise_loss}:
    \begin{align} \label{eq:denoise_loss}
         D^*_\theta =\arg\min_{D_\theta} \mathbb{E}_{p_{0,t}}[\|D_\theta(\bfX_{t},t) - \bfX_{0}\|^2].
    \end{align}
    \textbf{Conditional Generation}.
    Conditional generation is typically achieved via a conditional score, which can either be trained by DSM or decomposed into a unconditional and guidance term. The unconditional score can be pre-trained via DSM and the guidance term $\nabla \log p(y \mid x_t)$ , which can be approximated in many ways such as via some classifier \citep{dhariwal2021diffusion}; classifier on a denoised state \citep{chungdiffusion} or simply trained via denoising \citep{ho2022classifier, denker2024deft}:
    \begin{equation}
        \nabla \log p_t(x_t \mid  y) =  \nabla \log p_t(x_t) + \omega \nabla \log p_t(y \mid x_t).
    \end{equation}
    Here $\omega \geq 1$ heuristically adjusts guidance strength, similar to temperature controlled sampling.  Classifier-free guidance is the most commonly used training approach, estimating the guidance term by $\nabla \log p(y \mid x_t)=\nabla \log p(x_t \mid y) - \nabla \log p_t(x_t)$, where each individual term is trained via conditional DSM. 

\subsection{Energy Based Models}\label{sec:ebm}
Energy-based models (EBMs) \citep{lecun2006tutorial} approximate density $\pdata \approx p_\theta:= Z^{-1}e^{-E_\theta}$ using $\theta$ parameterized potential $E_\theta(x)$, for normalising constant $Z$. The seminal work of~\citet{teh2003energy}, later improved by~\citet{du2021improved}, introduced contrastive training of EBM by taking gradient steps on $-\expect_{x\sim p_0}[E_{\theta}(x)]+\expect_{x\sim p_\theta}[E_{\theta}(x)]$. Sampling $p_\theta$ typically entails expensive MCMC methods however.  

\textbf{Diffusion models as a sequence of energy based models}.
Given the idealised score, $\nabla \log p_t$, is a gradient one could avoid MCMC and use denoising score matching to learn a sequence of energy functions $(E_\theta(\cdot,t))_t$ such that $-\nabla_{x_t} E_\theta(x_t,t) \approx \nabla \log p_t(x_t)$, i.e. $s_\theta(x_t,t):=-\nabla_{x_t} E_\theta(x_t,t)$, henceforth referred to as an energy-parameterisation. This is in contrast to the usual diffuson model parameterisation where score or denoiser is approximated directly with a neural network $s_\theta(x_t,t) \approx \nabla \log p_t(x_t)$. Energy parameterized diffusion models were first shown to be possible for image datasets in \cite{salimans2021should} by careful choice of architecture, yet thus far remains noticeably inferior to unconstrained diffusion models.

\subsection{Sequential Monte Carlo} 
Before delving into our method, we briefly recap Sequential Monte Carlo (SMC)  \citep{doucet2001sequential, chopin2020introduction} for later use in \Cref{sec:comp_cont}. SMC entails propagating $K$ particles initially sampled from some distribution $M_0$ through a sequence of proposal, importance weighting, and resampling steps. The resampling steps are crucial to ensuring computation is focused on promising particles, and to avoiding weight degeneracy. A simplified algorithm is presented in \Cref{alg:gen_smc}, any resampling approaches could be used, in practice we use adaptive resampling \citep{del2011adaptive} with the systematic resampler \citep[Chapter 4]{chopin2020introduction}.

SMC enables an approximate change of measure through the Feynman Kac model (FKM) framework \citep{chopin2020introduction, del2004feynman}. A FKM consists of an initial distribution $M_0$; some time indexed Markov transition kernels $(M_t)_t$, which we can sample from; and non-negative potential functions $(G_t)_t$, $G_0: \mathcal{X} \rightarrow \mathbb{R}^+$, $G_t: \mathcal{X}^2 \rightarrow \mathbb{R}^+$. 
\begin{align}
    &M(\mathrm{d}x_{0:T}) = M_0(\mathrm{d}x_0)\prod_{t=1}^{T}M_t(\mathrm{d}x_t|x_{t-1}) \label{eq:proposal_measure} \\ \vspace{-0.1cm}
    &Q(\mathrm{d}x_{0:T}) \propto G_0(x_0)\prod_{t=1}^{T}G_t(x_t,x_{t-1})M(\mathrm{d}x_{0:T}) \label{eq:fk_measure} 
\end{align}
The use of potentials permits a change of measure from the proposal from Markov process \eqref{eq:proposal_measure} to the FKM distribution \eqref{eq:fk_measure}, where \eqref{eq:fk_measure} can be approximately simulated with SMC or particle filtering as in \Cref{alg:gen_smc}.
{\small
\begin{algorithm}
        \caption{Generative SMC}
        \label{alg:gen_smc}
        \setlength{\parindent}{0pt}
        \begin{algorithmic}
            \State{Sample $\bfX_{0}^{k}\stackrel{\text{i.i.d.}}{\sim} M_0$ ~for $k \in [K]$}
           \State{Weight $\omega_1^{k}=G_0(\bfX^{k}_0)$~for $k \in [K]$}
        \For{$t=1,...,T$}
            \State{Normalize weights $w^{k}_{t-1}\propto \omega^{k}_{t-1}$, $\sum_{k=1}^K  w^{k}_{t-1}=1$}
            \State{Resample $\tilde{\bfX}^{k}_{t-1}\sim \sum_{k=1}^K  w^{k}_{t-1} \delta_{\bfX^{k}_{t-1}}$ ~for $k \in [K]$}
                \State{Proposal $\bfX_{t}^{k}\sim M(\cdot|\tilde{\bfX}^{k}_{t-1})$  ~for $k \in [K]$}
                \State{Weight $\omega^{k}_t= G(\bfX_{t}^{k},\tilde{\bfX}^{k}_{t-1})$}
        \EndFor
        \State{{\bfseries Return:} samples $(\bfX_{T}^{k})_k$}
        \end{algorithmic}
        \end{algorithm}
        }

\begin{figure*}[!hb]
  \setlength\tabcolsep{6pt}
  \adjustboxset{width=\linewidth,valign=c}
  \centering
  \begin{tabularx}{1.0\linewidth}{@{}
      l
      X @{\hspace{6pt}}
      X
      X
      X
      X
      X
    @{}}
    \rotatebox[origin=c]{90}{\textbf{Ours}}
    & \includegraphics{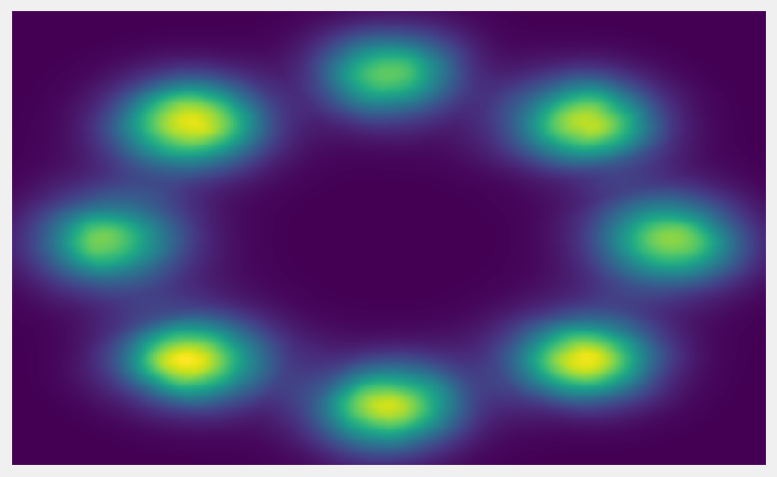}
    & \includegraphics{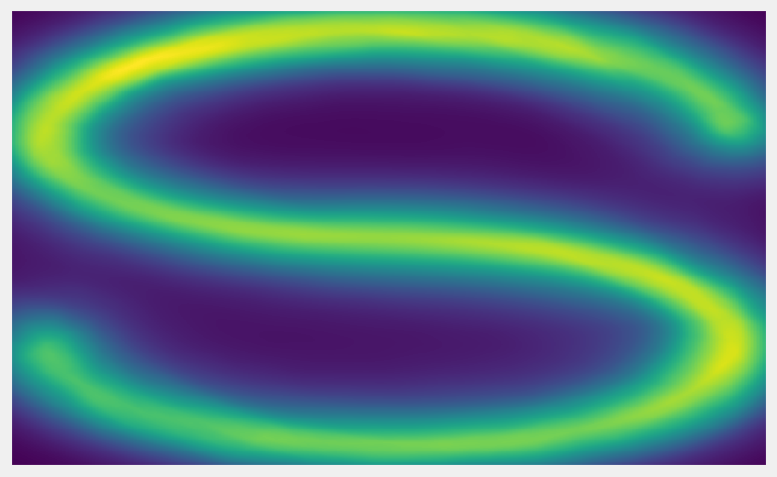} 
    & \includegraphics{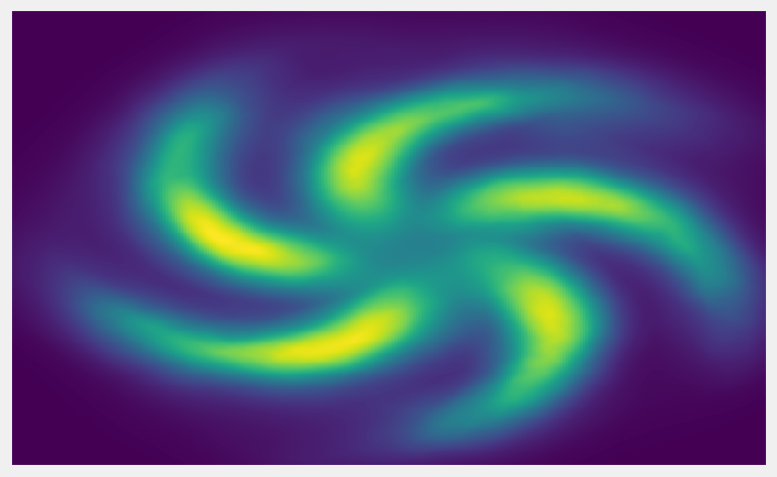} 
    & \includegraphics{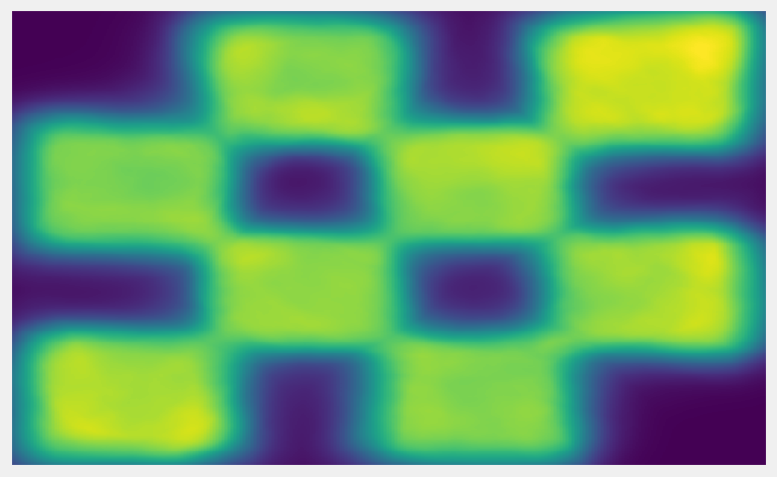} 
    & \includegraphics{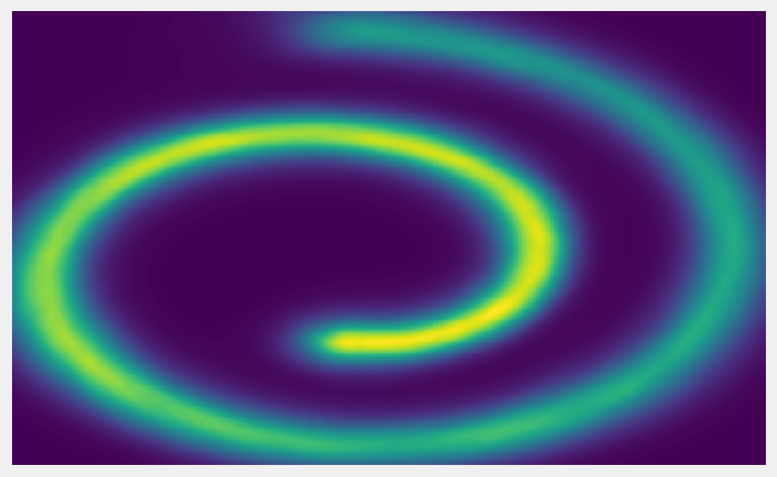}
    & \includegraphics{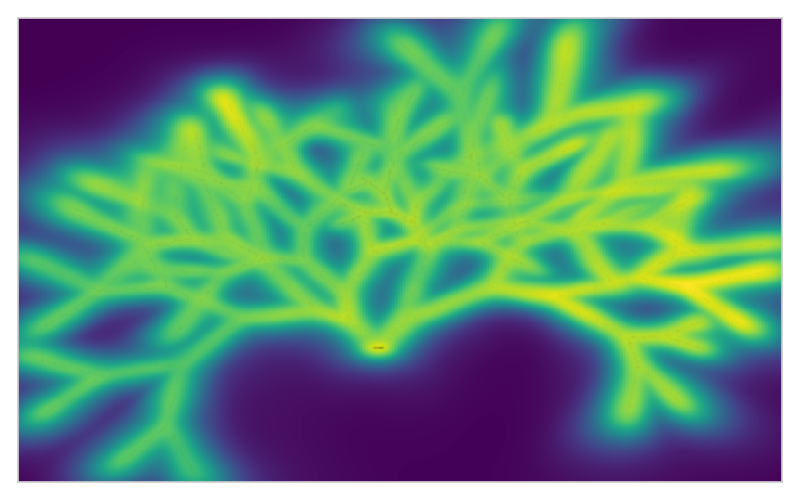} \\
    \rotatebox[origin=c]{90}{\textbf{E-DSM}}
    & \includegraphics{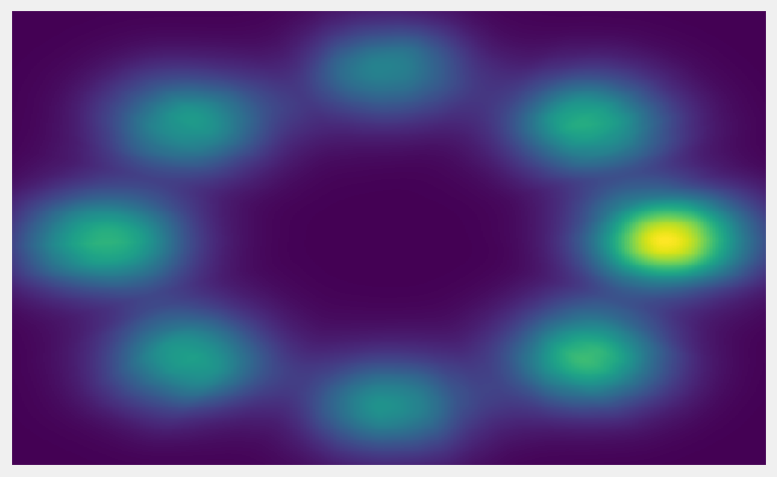}
    & \includegraphics{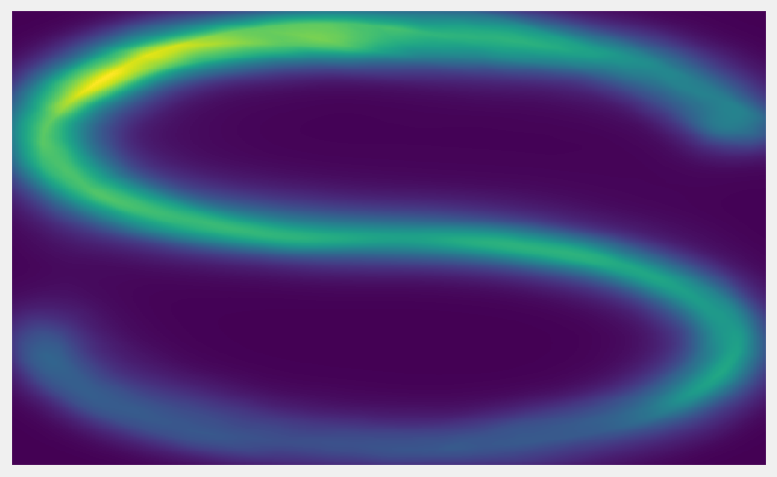}
    & \includegraphics{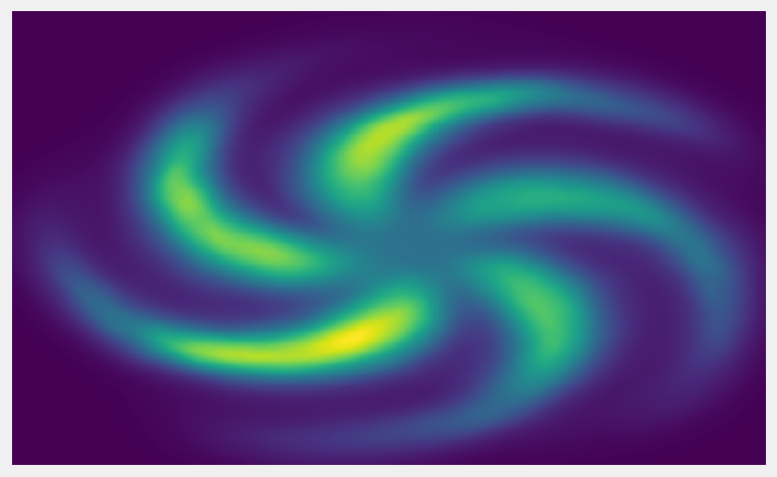}
    & \includegraphics{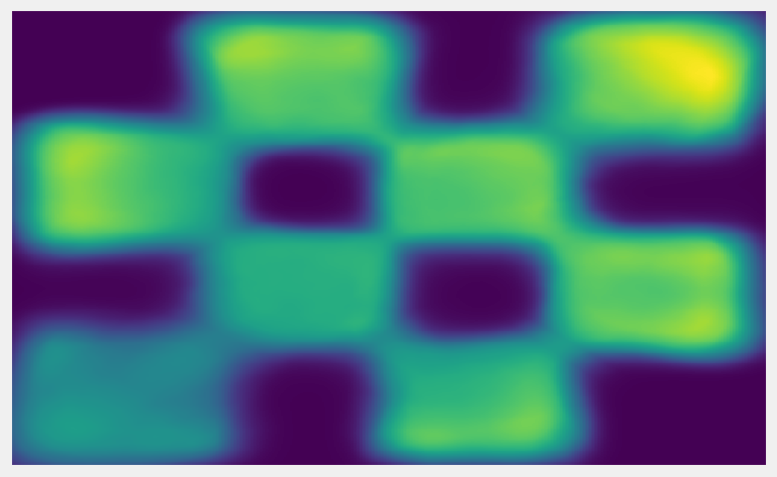}
    & \includegraphics{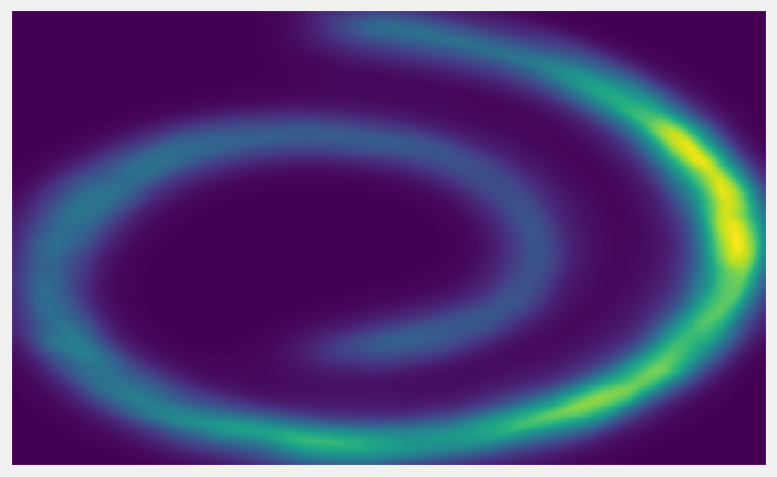}
    & \includegraphics{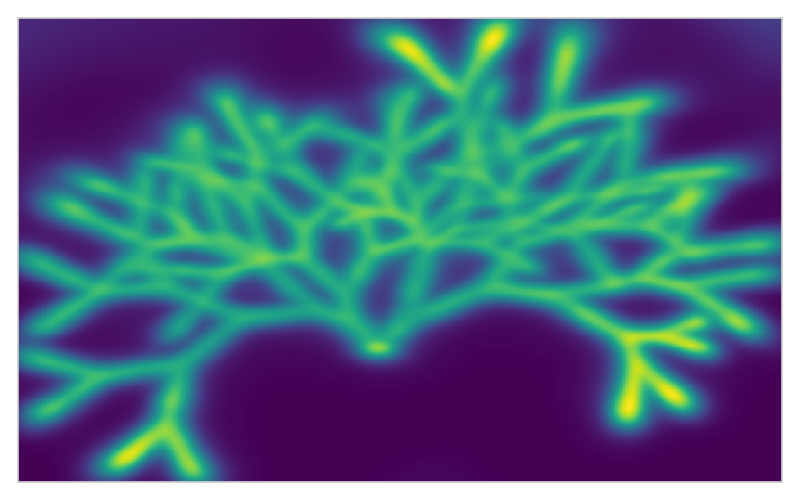} 
  \end{tabularx}
  \caption{Density plot of $p_E\propto\exp{-E_\theta(x_t,t)}$, where $E_\theta$ is trained via distillation of pre-trained diffusion networks (Ours) (top) vs via energy parameterized denoising score matching (E-DSM).}
  \label{fig:2d_density}
\end{figure*}
\newpage
\section{Distilled Energy Diffusion Models} \label{sec:training_methods}
\subsection{Training Instability with Energy Parameterised Diffusion Models}

Denoising score-matching \citep{vincent2011connection, song2019generative} is a promising simulation-free alternative to contrastive learning for training energy based models \citep{salimans2021should, song2021trainebm}. We call DSM with an energy-parameterised score E-DSM. Although E-DSM never quite learns exactly the energy at time $t=0$; it can approximate the energy arbitrarily close and been used successfully in generative modelling \citep{du2023reduce, salimans2021should} and sampling \citep{phillips2024particle}.

\begin{figure}[H]
    \centering
        \includegraphics[trim={0 0.cm 0 0},clip,width=\linewidth]{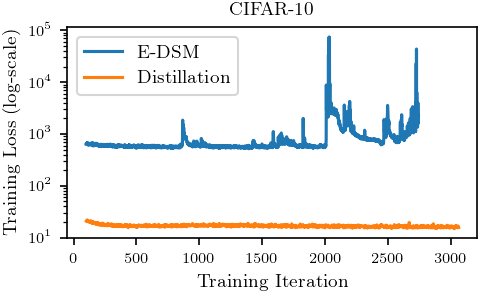}
    \caption{E-DSM loss (blue): $\mathbb{E}\|D_\theta(\bfX_t,t)- \bfX_0\|$ without gradient clipping vs distillation loss (orange): $\mathbb{E}\|D_\theta(\bfX_t,t)- D^{teach}_{\phi}(\bfX_{t},t)\|$ during training of a diffusion model for CIFAR10. Initial $100$ iterations cut.} 
    \label{fig:loss_stability}
\end{figure}

Unfortunately, E-DSM suffers from training instability, as shown in \Cref{fig:loss_stability}. We attribute this instability due two effects: firstly that DSM has a high variance loss \citep{jeha2024variance}, and secondly network architecture limitations. 
Unlike in regular DSM or contrastive training,  E-DSM requires taking gradients of $E_\theta$ with respect to both the state and parameters i.e. $\nabla_\theta\nabla_x E_\theta(x,t)$ during training. The effective network in E-DSM, $-\nabla_x E_\theta(x, t)$, may be viewed informally as the composition of $E_\theta: \mathbb{R}^d \rightarrow \mathbb{R}$ and operation $\nabla_x: \mathbb{\rset} \rightarrow \rset^d$. The second operation, $\nabla_x$ lacks any normalisation or residual connections common in modern unconstrained neural networks typically used for diffusion models. Such stability measures have been shown to be crucial \citep{Karras2024edm2, karras2022elucidating} to ensuring stable training and generative performance.

We tackle both weaknesses by first providing a more stable loss, and secondly with careful parameterisation and initialization of the energy-diffusion network.

\subsection{Distillation Loss}
We first introduce the conservative projection loss as: 
\begin{align} \label{eq:conservative_proj}
    \arg\min_{\theta} \mathbb{E}_{p_{t}}[\|\nabla u_\theta(\bfX_{t},t) - v(\bfX_{t},t)\|^2],
\end{align}
where $u_\theta$ is some flexibly parameterised function, such as a neural network. A conservative vector field is the gradient of some potential. The loss \eqref{eq:conservative_proj} aims to learn a potential $u_\theta$ and hence conservative vector field $\nabla u_\theta$ which is closest in squared Euclidean distance to $v$, not necessarily conservative.

Consider ODEs generated by $v$ and the minimizer of \eqref{eq:conservative_proj}, $u_{\theta^*}$:
\begin{align}
    \mathrm{d}\bfX_t &= v(\bfX_t,t)\mathrm{d}t & 
    \mathrm{d}\bfX_t &= u_{\theta^*}(\bfX_t,t)\mathrm{d}t. \label{eq:given_vf}
\end{align}
Directly using \cite[Theorem 5.2]{liu2022rectified}, if $v$ is locally bounded, and \eqref{eq:given_vf}(left) has a unique solution generating density $p_t$, then the marginals of ODEs in \eqref{eq:given_vf} coincide. Minimising \eqref{eq:conservative_proj} may therefore be considered a type of Helmholtz decomposition of $v$, where the ``rotation-only'' component of $v$ is removed, discussed further in \Cref{app:conservative}.

Note: a more general class of Bregman Helmoholtz losses have been considered in the seminal work of \citet{liu2022rectified} in the context of rectified flows.

\textbf{Distilling a Score into an Energy}.
In the context of training energy based models, we consider $u_\theta = -E_\theta$. Ideally we would use $v=\nabla \log p_t$, in such a case \eqref{eq:conservative_proj} would simply be (non-denoising) score-matching \citep{hyvarinen05a}. We do not have access to $\nabla \log p_t$ so instead use a pre-trained score-function $s^{teach}_{\phi}$, i.e. $v(\bfX_{t},t)=s^{teach}_{\phi}$ as proxy:
\begin{align} \label{eq:score_energy_distill}
    \arg\min_{\theta} \mathbb{E}_{p_{0,t}}[\|\nabla E_\theta(\bfX_{t},t) + s^{teach}_\phi(\bfX_{t},t)\|^2].
\end{align}
By the arguments above, we do not require $s^{teach}_{\phi}$ be conservative, as the minimizer of \eqref{eq:score_energy_distill} will generate the same distribution as $s^{teach}_{\phi}$ via the probability flow ODE and hence generate a distribution close to the data distribution, if $s^{teach}_{\phi}$ is well trained.

 \textbf{Expressed as Denoising}. The loss \eqref{eq:score_energy_distill} may equivalently be written as a denoising loss with target $D^{teach}_{\phi}(x_{t},t) \approx \mathbf{E}[\bfX_0|x_t]$, as in \eqref{eq:denoising_distill} where again by Tweedie's formula, one may write $s^{teach}_{\phi}(x{t},t)$ in terms of denoiser $D^{teach}_{\phi}(x_{t},t)=\alpha_t^{-1}[x_t+\sigma_t^2 s^{teach}_{\phi}(x_{t},t)]$. 
\begin{align} \label{eq:denoising_distill}
    \arg\min_{D_\theta} \mathbb{E}_{p_{0,t}}[\| D_\theta(\bfX_{t},t) - D^{teach}_{\phi}(\bfX_{t},t)\|^2].
\end{align}
The corresponding distilled denoiser is related to the energy through $D_\theta(x_t,t)= \alpha_t^{-1}[x_t\sigma_t^2 -\nabla E_{\theta}(x_{t},t)]$.

Losses \eqref{eq:denoising_distill} and \eqref{eq:denoise_loss} differ only in the regression target: $D^{teach}_{\phi}(\bfX,t) \approx \mathbf{E}[\bfX_0|x_t]$ vs $\bfX_0$ respectively. Given $\mathbf{E}[\bfX_0|x_t]$ is the minimizer of \eqref{eq:denoise_loss} and at large time $t$, $\mathbf{E}[\bfX_0|x_t]$ is quite different to $\bfX_0$, and hence intuitively targeting $D^{teach}_{\phi}(\bfX,t) \approx \mathbf{E}[\bfX_0|x_t]$ directly results in more stable training than \eqref{eq:denoise_loss}.

 Motivated by the desire to reduce training instability, we have framed this distillation loss in terms of learning energy-parameterised diffusion models; however, the same loss could also be applied to train unconstrained diffusion models through distillation.

\begin{figure*}[hb]
    \centering
    \begin{subfigure}[t]{0.19\textwidth}
        \centering
        \includegraphics[width=\linewidth]{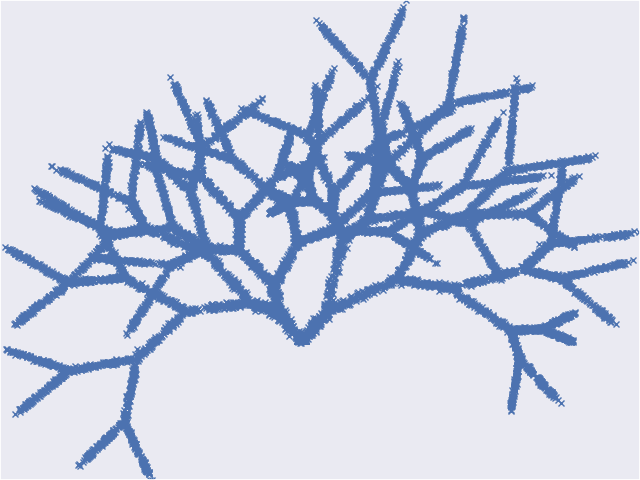}
        \caption{Ground Truth}
    \end{subfigure}%
    ~ 
    \begin{subfigure}[t]{0.19\textwidth}
        \centering
        \includegraphics[width=\linewidth]{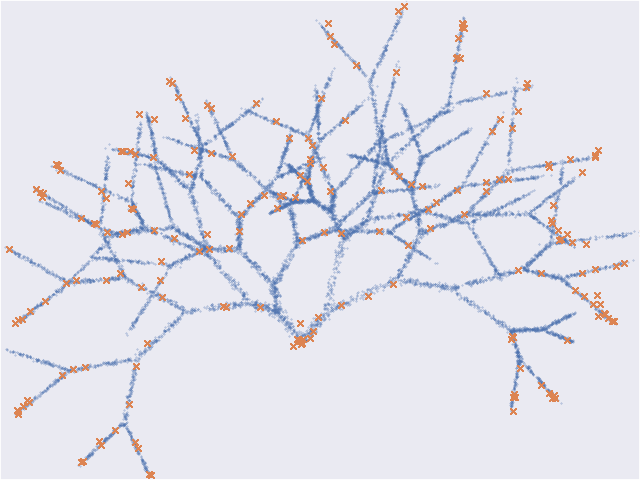}
        \caption{$\gamma=-0.05$}
    \end{subfigure}%
    ~ 
    \begin{subfigure}[t]{0.19\textwidth}
        \centering
        \includegraphics[width=\linewidth]{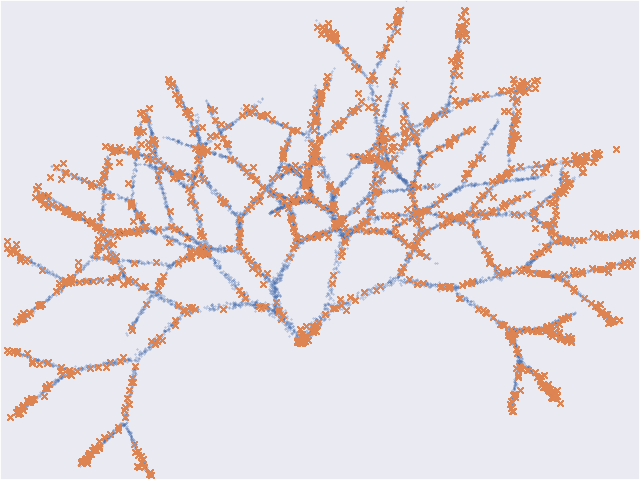}
        \caption{$\gamma=-0.04$}
    \end{subfigure}%
    ~ 
    \begin{subfigure}[t]{0.19\textwidth}
        \centering
        \includegraphics[width=\linewidth]{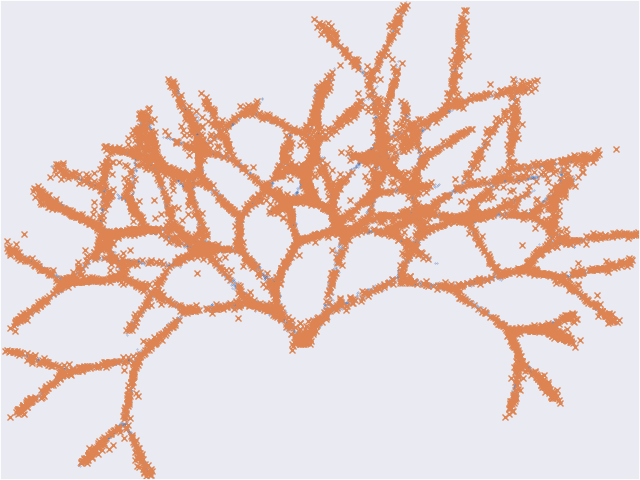}
        \caption{$\gamma=0.0$}
    \end{subfigure}%
    ~ 
    \begin{subfigure}[t]{0.19\textwidth}
        \centering
        \includegraphics[width=\linewidth]{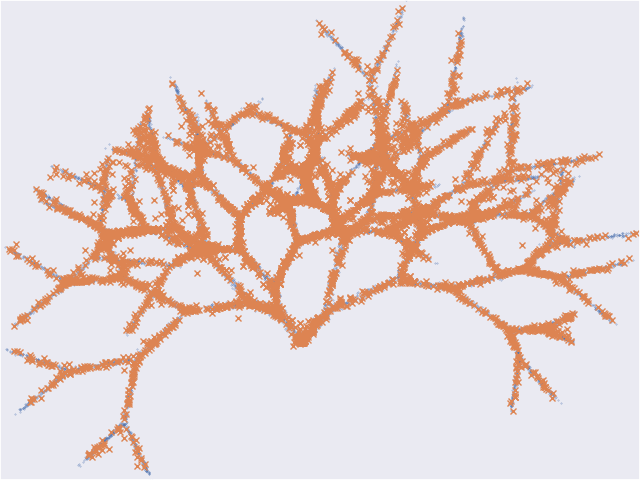}
        \caption{$\gamma=0.05$}
    \end{subfigure}
    \\
    \begin{subfigure}[t]{0.19\textwidth}
        \centering
        \includegraphics[width=\linewidth]{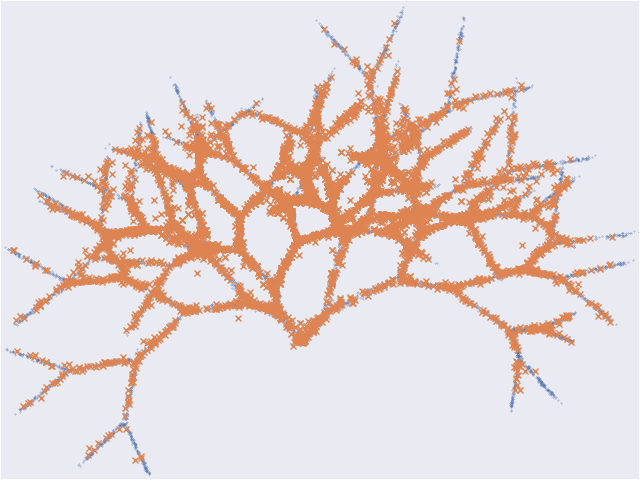}
        \caption{$\gamma=0.1$}
    \end{subfigure}%
    ~ 
    \begin{subfigure}[t]{0.19\textwidth}
        \centering
        \includegraphics[width=\linewidth]{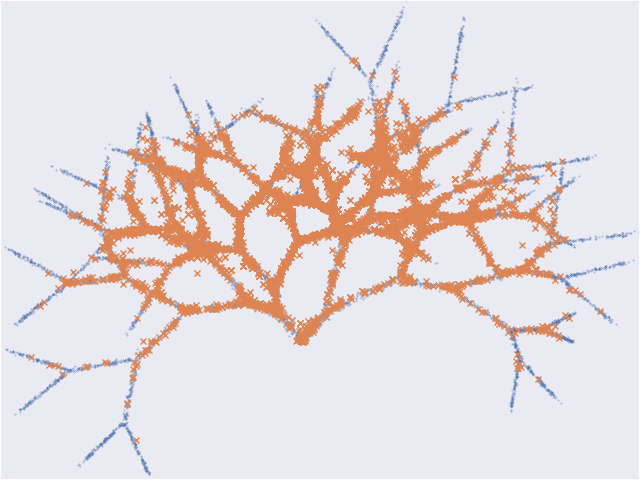}
        \caption{$\gamma=0.2$}
    \end{subfigure}%
    ~ 
    \begin{subfigure}[t]{0.19\textwidth}
        \centering
        \includegraphics[width=\linewidth]{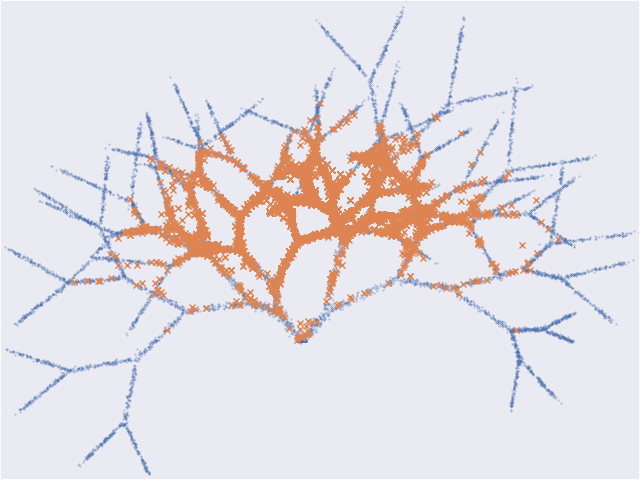}
        \caption{$\gamma=0.5$}
    \end{subfigure}%
    ~ 
    \begin{subfigure}[t]{0.19\textwidth}
        \centering
        \includegraphics[width=\linewidth]{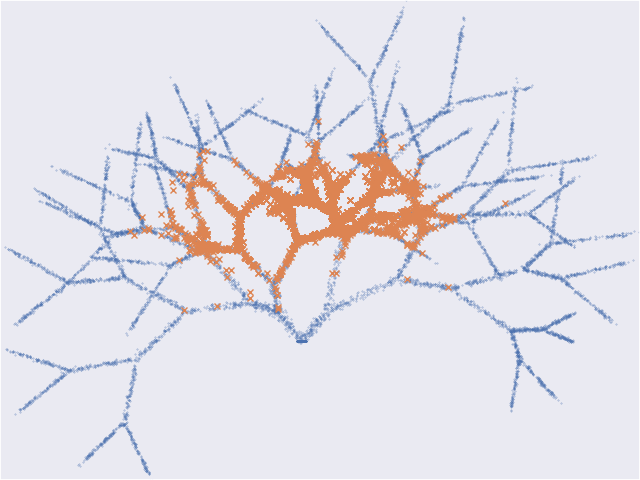}
        \caption{$\gamma=1.0$}
    \end{subfigure}%
    ~ 
    \begin{subfigure}[t]{0.19\textwidth}
        \centering
        \includegraphics[width=\linewidth]{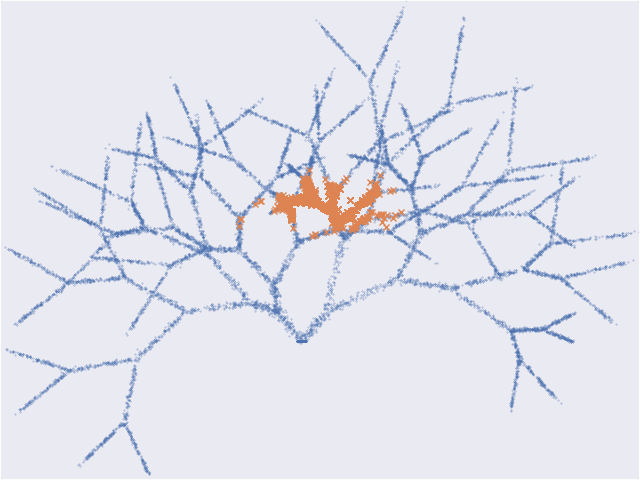}
        \caption{$\gamma=5.0$}
    \end{subfigure}%
    \caption{\textbf{SMC Sampling of Feynman Kac Diffusion Models} for $G_i(x_{t_{i}},x_{t-1})=\exp\{-\gamma_{t_i} E_\theta(x_{t_i}\}\approx p(x_{t_i})^\gamma_{t_i}$. The fractal distribution, inspired by~\citet{karras2024guiding}, is obtained by fitting Gaussian mixtures to each branch and appending recursively. Ground truth samples shown in (faded) blue and generated samples shown in orange.}   
    \label{fig:temp_tree}
\end{figure*}
\begin{figure}[H]
    \centering
        \includegraphics[trim={0 0.1cm 0 0},clip,width=\linewidth]{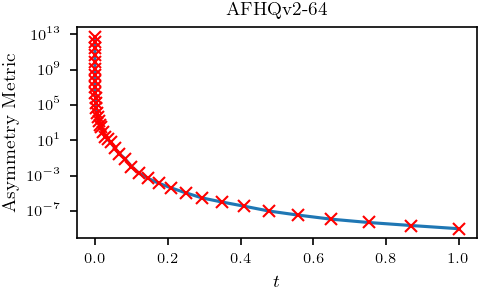}
    \caption{Asymmetry metric in log-scale, $\|\jac - \jac^T\|^2$ where Jacobian $\jac=\mathbf{D}_x s_\theta(x_t,t)$ of score network $s_\theta$ trained via DSM on AFHQv2-$64$, $t\in[0,1]$.} 
    \label{fig:afhq_asymmtery}
\end{figure}

\textbf{Scores are approximately conservative}.
Similar to \citet{lai2023fp}, we observe that well-trained score networks are approximately conservative, except at close to $t=0$, where the score exhibits a Lipschitz singularity~\citep{yang2023lipschitz}, see \Cref{fig:afhq_asymmtery}. Here \textit{conservativity} of a score network may be quantified by the asymmetry of its Jacobian (see \Cref{app:conservative}).

\newpage
\subsection{Parameterization and Initialization}
\textbf{Pre-conditioning}.
To further reduce training instability of energy-parameterised diffusion models; we use the preconditioning ( $c_\text{s}, c_{out}, c_\text{in}, c_\text{t}$) of \citet{karras2022elucidating}, parameterizing the denoiser, $D_\theta$ as:
\begin{align*}
    D_\theta(x_t,t) &= c_\text{s}(t)x_t + c_{out}(t)\nabla_x F_\theta(c_\text{in}(t)x_t,c_\text{t}(t)).
\end{align*} 
We replace the unconstrained network in \cite{karras2022elucidating} with the gradient of network $F_\theta$. By Tweedie \citep{efron_tweedie}, we compute the energy via:
\begin{align*}
    E_\theta(x_t,t) &= \tfrac{1-\alpha_t c_\text{s}(t)}{2\sigma_t^{2}}\|x_t\|^2
    -\tfrac{\alpha_t c_{out}(t)}{c_\text{in}(t)\sigma_t^{2}}F_\theta(c_\text{in}(t)x_t,c_\text{t}(t)).
\end{align*}
\textbf{Network parameterization}. We parameterize the network $F_\theta$ such that its gradient takes a similar network structure to score/ denoising networks. Consider a network architecture, $h_\theta$, known to work well for DSM, such as a U-Net for image-based diffusion models. The following parameterization forces $\nabla_x F_\theta$ to resemble $h_\theta(x_t,t)$ with the addition of a residual term, where $\mathbf{D}_x h_\theta$ denotes Jacobian of $h_\theta$:
\begin{align} \label{eq:parameterisation}
    F_\theta(x_t,t) &= h_\theta(x_t,t)\cdot x_t\\
    \nabla_x F_\theta(x_t,t) &= x_t \cdot \mathbf{D}_x h_\theta(x_t,t)+h_\theta(x_t,t)
\end{align}
This structure is important for initialization. 

\textbf{Network initialization}. Motivated by recent successes in initializing models with pre-trained diffusions \citep{lee2024improving, kim2024simple} we do the same for our energy-parameterized network. In particular, we set  $h_\theta$ to be the same architecture as the teacher network $D^{teach}_{\phi}$ and initialize with teacher parameters $\theta \gets \phi$. This simple technique is effective due to choice in parameterization \eqref{eq:parameterisation}, and results in drastically faster convergence (see \Cref{app:experiments}) and better generative performance overall.

\begin{figure*}[!t]
   \centering \includegraphics[trim={0 13cm 0 2.3cm},clip,width=\linewidth]{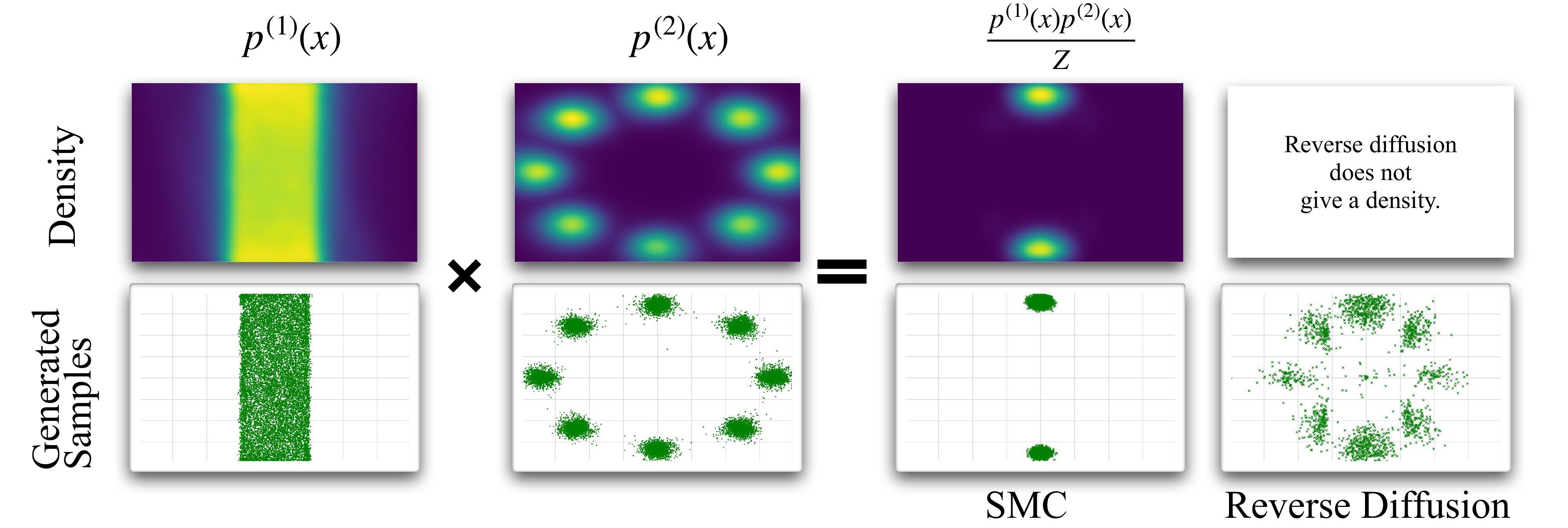} 
 \caption{Simple 2D Composition failure from \cite{du2023reduce}. Top row: Learnt densities, $e^{-E^{(i)}_\theta}$ for each $p^{(i)}_{t}$ and $e^{-E^{(1)}_\theta-E^{(2)}_\theta}$. Bottom row: generated samples per $p^{(i)}_{t}$, as well as the SMC generation using \eqref{eq:comp_fkm} and reverse diffusion of summed scores,  ${q}_{\theta, \lambda}^{(1)+(2)}(x_{t_{0:N}})$.}
 \label{fig:toy_composition}
\end{figure*}
\newpage
\section{Composition and Control} \label{sec:comp_cont}
\subsection{Feynman Kac Diffusion Models}
As discussed in \Cref{sec:background}, the Feynman Kac model (FKM) formalism \citep{chopin2020introduction, del2004feynman} is a simple yet principled framework to perform an approximate change of measure from a given Markov process according to a user-provided potentials. Given diffusion models are expensive to train, it is desirable to use pretrained models. 

Denote the discretization of the generative diffusion process \eqref{eq:time_reversed} as $q^\lambda(x_{t_{0:N}})$, and the network approximation as $q_\theta^\lambda(x_{t_{0:N}})$ in \eqref{eq:diff_markov} for any ODE/ SDE solver on discretisation {${T=t_0 \geq t_1 \geq \ldots \geq t_N=0}$}. We choose the underlying Markov measure for a FKM to be $q^\lambda_\theta$ or similarly, some conditioned process $q^\lambda_\theta(\cdot|y)$ for conditioning signal $y$ e.g. labels. Explicit forms of ${q}_\theta^\lambda(x_{t_{i+1}}|x_{t_{i}})$ are given in \Cref{app:diffusion_samplers}.
\vspace{-0.1cm}
\begin{align}\label{eq:diff_markov}
    &{q}_\theta^\lambda(x_{t_{0:N}}) = q^\lambda_{t_0}(x_{t_0})\prod_{i=0}^N {q}_\theta^\lambda(x_{t_{i+1}}|x_{t_{i}}) 
\end{align}
The intermediate FKM distributions by running \Cref{alg:gen_smc} with potentials $(G_i)_i$ and Markov process \eqref{eq:diff_markov} for $n\leq N$ are then given by:
\vspace{-0.3cm}
\begin{align}\label{eq:fkm_diff}
    &Q(x_{t_{0:n}}) \propto {q}_\theta^\lambda(x_{t_{0:n}})G_0(x_{t_0})\prod_{i=1}^{n}G_i(x_{t_i},x_{t_{i-1}})
\end{align}
\vspace{-0.5cm}
\subsection{Temperature-Controlled Generation}
Let $\gamma_t \in \mathbb{R}$ be some time-indexed inverse temperature parameter. Given access to density $p_{t_i}$, one could set $G_i(x_{t_i}, x_{t_{i-1}})=p(x_{t_i})^{\gamma_{t_i}}$. Rearranging \eqref{eq:diff_markov} gives:
\begin{align*}
    Q(x_{t_{0:N}}) &\propto \prod_{i=0}^{N}p(x_{t_i})^{\gamma_{t_i}}q^\lambda(x_{t_{0:N}}). 
\end{align*}
Recall from \Cref{sec:background}, regardless of choice of $\lambda$, the marginals of the backward and forward process match, $q^\lambda(x_{t})=p(x_{t})$ for any $t$, hence:
\begin{align*}
    Q(x_{t_{0:N}}) &\propto p(x_{t_N})^{1+\gamma}q^\lambda(x_{t_{0:N-1}}|x_{t_N})\prod_{i=0}^{N-1}p(x_{t_i})^\gamma  \label{eq:diff_fk_measure} 
\end{align*}
\Cref{fig:temp_tree} illustrates how setting $\gamma \geq 0$ results a more concentrated distribution, $p(x_{t_N})^{1+\gamma}$. Similarly,  $\gamma<0$ results lower density regions being sampled with greater probability. This is biased toward rare events.  

In practice, we substitute $G_i(x_{t_{i}},x_{t-1}) \gets e^{-\gamma_{t_i} E_\theta(x_{t_i})}$ as an approximation, proportional to $p(x_{t_i})^\gamma_{t_i}$. 

It is common to proxy low temp. generation with high guidance weights, as detailed in \Cref{sec:background}. Our SMC approach provides an alternative which may be used with unconditional, conditional models. An example demonstrating low-temperature generation is shown in \Cref{exp:low_temp} for conditional image generation. 

\subsection{Compositional Generation}\label{sec:composition_scores}
EBMs enable composition of pretrained models  as logical operators can be expressed as functions of the score and energy \citep{compose_ebm, compose_diffusion}, see \Cref{app:composition}. We focus here on the \textsc{AND} operation. 

Unlike for EBMs the summed scores from diffusion models at $t>0$ do not always match the score for the composed distribution \citep{du2023reduce}. Consider time indexed densities $p^{(1)}_{t}$ and $p^{(2)}_{t}$, and denote the composed score:
\begin{equation}\label{eq:compose_score}
    s^{(1)+(2)}_t(x_t, t) = \nabla \log p^{(1)}_{t}(x_t) + \nabla \log p^{(2)}_{t}(x_t)
\end{equation}
Let ${q}_{\lambda}^{(1)+(2)}(x_{t_{0:N}})$ denote the process \eqref{eq:time_reversed} with composed score \eqref{eq:compose_score}, and consider FKM given by \eqref{eq:comp_fkm}.
\begin{align}
    &M_t(x_{t_{i+1}}|x_{t_{i}})= q^\lambda_{t_0}(x_{t_0})\prod_{i=0}^N {q}_{\lambda}^{(1)+(2)}(x_{t_{i+1}}|x_{t_{i}}) \label{eq:comp_fkm}\\ &M_t(x_{t_{i+1}}|x_{t_{i}})G_i(x_{t_{i+1}}, x_{t_{i}}) = p^{(1)}_{t}(x_{t_{i+1}})p^{(2)}_{t}(x_{t_{i+1}}). \notag
\end{align}
In practice we use the approximation ${q}_{\theta, \lambda}^{(1)+(2)}(x_{t_{0:N}})$ defined by summing the trained score functions for $p^{(1)}_{t}$ and $p^{(2)}_{t}$, and similarly approximate, up to a scalar, each $p^{(i)}_{t}$ with $e^{-E^{(i)}_\theta(\cdot,t)}$. Expanding the FKM intermediate distributions for this choice of $G$ in \eqref{eq:fkm_diff} gives a sequence of product densities coinciding with the target density we wish to generate. A simple example of this is illustrated in \Cref{fig:toy_composition}, note that the density is given by the energy function, and hence reverse diffusion does not provide a density directly.

\subsection{Bounded Generation}\label{sec:bounded} 
We consider how constraints \citep{lou2023reflected, fishman2024metropolis, fishmandiffusion}  and dynamic thresholding \citep{saharia2022photorealistic} can be imposed via FKM potentials. \cite{fishman2024metropolis} adjusts sampling by rejecting transitions if proposals $x_{t_i}$ fall outside a specified region, $B$. This resembles a FKM with potential $G_i(x_{t_i}, x_{t_{i-1}})= \mathbb{I}_B(x_{t_i})$. Similarly, dynamic thresholding entails clipping the denoiser to be within a unit-cube at generation time. This also resembles a FKM setting $G_i(x_{t_i}, x_{t_{i-1}})= \mathbb{I}_{[-1+\delta,1-\delta]^d}(D_\theta(x_t,t))$, $\delta>0$.

The above choices of regions are simple, but quite crude. One could instead use the energy as a softer alternative, for example by choosing $G_i(x_{t_i}, x_{t_{i-1}})=\exp\{-\gamma_tE_\theta(x_t,t)\}$ or $G_i(x_{t_i}, x_{t_{i-1}})=\exp\{-\gamma_tE_\theta(D_\theta(x_t,t),\epsilon)\}$, $\epsilon\approx0$. See generated examples in \Cref{app:fkm_potentials}.

\begin{figure}[b]
    \centering
    \begin{subfigure}[t]{0.45\linewidth}
        \centering
        \includegraphics[width=\linewidth]{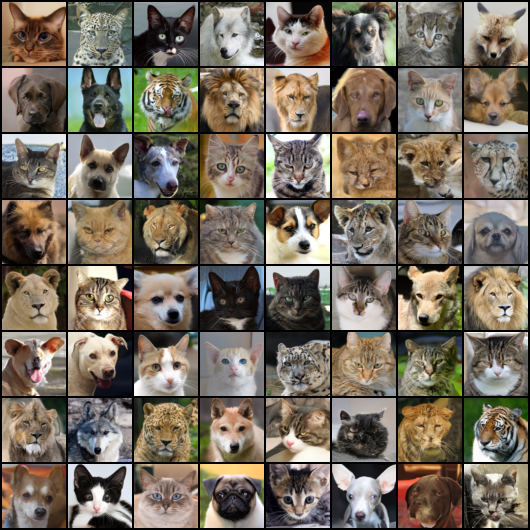}
        \caption{Distilled, Ours}
    \end{subfigure}%
    ~
    \begin{subfigure}[t]{0.45\linewidth}
        \centering
        \includegraphics[width=\linewidth]{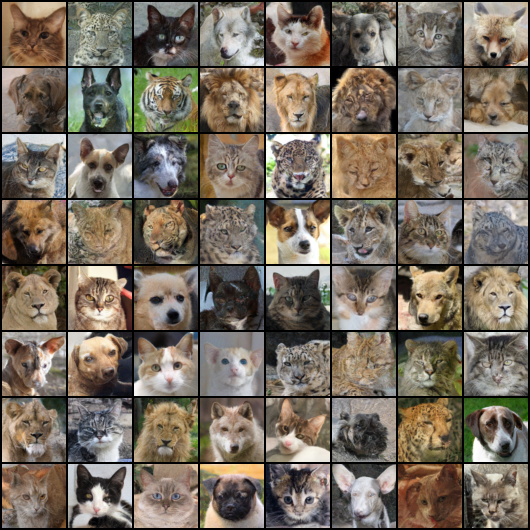}
        \caption{E-DSM}
    \end{subfigure}%
    \caption{Samples from AFHQv2-$64$ using energy parameterized diffusion models trained with our distillation method vs denoising score-matching (E-DSM)} 
    \label{fig:afhq}
\end{figure}
\section{Experiments}
Full experimental details including network architectures and training recipes are provided in \Cref{app:experiments}.
\subsection{2D Experiments}

\textbf{E-DSM vs Distillation}. \Cref{fig:2d_density} provides qualitative comparison of our distillation approach compared to E-DSM, We used a feedforward network with sine nonlinearity. There is a general uneven density exhibited within E-DSM, not present in our approach.

\textbf{Temperature Controlled Generation.} Figure~\ref{fig:temp_tree} illustrates the effects of temperature on the fractal dataset. We are able to generate either the high density regions in low temperature regime, and to target low density regions (at the boundary of the support) with higher temperatures.   

\textbf{Compositional Generation.} \Cref{fig:toy_composition} demonstrates the failure of reverse diffusion in composition. Our SMC based composition approach however correctly recovers the joint distribution $p_1(x)p_2(x)$. A similar result has been observed in \citet{du2023reduce}, where they correct with MCMC rather than SMC.

\subsection{Generative Performance of Diffusion-Energy Models}
We first verify our improvements for training energy-parameterized diffusion models in terms of generative performance, measured by Frechet Inception Distance (FID) \citep{heusel2017gans} on standard EBM datasets: CIFAR-10 \citep{cifar} and CelebA-64 \citep{celeba} for both unconditional generation in \Cref{tab:unconditional_gen} and conditional generation in \Cref{tab:conditional}. In the interest of transparency, we display the Number of Function Evaluations (NFE) involved in the generative process. We further showcase our method with AFHQv2-64 \citep{afhq}, FFHQ-64 \citep{ffhq} in \Cref{tab:afhq} and visually in \Cref{fig:afhq}. We compare to recent energy-parameterized approaches \citep{gao2020learning, zhu2023learning, hill2023tackling, schroder2024energy}, E-DSM, and popular methods such as diffusion \citep{karras2022elucidating} and flow-matching \citep{lipman2022flow, liu2022rectified}. Note: here we do not use SMC but standard generation with the gradient of the learnt energy.

Our method exhibits significantly better performance than other baselines using an energy-parameterization, particularly for CIFAR10 and AFHQv2. Although we achieve only modest performance improvement verses E-DSM on face-datasets, we note that the E-DSM results took significantly longer to train and required careful selection of learning rate and gradient clipping for stability, see \Cref{app:experiments}.
\begin{table}[ht]
\caption{Unconditional performance by NFE and FID for CIFAR-10 and CelebA. * denotes our result.EDM was used for the teacher score.} \label{tab:unconditional_gen}
\begin{center}
\begin{tabular}{lcccc}
\toprule
\multirow{2}{*}{\textbf{Method}} & \multicolumn{2}{c}{\textbf{CIFAR-$10$}} & \multicolumn{2}{c}{\textbf{CelebA-$64$}} \\
 & NFE & FID   & NFE & FID\\
\cmidrule{2-3} \cmidrule{4-5}
\textbf{Diffusion / Flow} \\
\begin{tabular}[c]{@{}l@{}}EDM* \\ {\scriptsize \citet{karras2022elucidating}}\end{tabular} & $35$   & $2.21$   & $79$  & $1.89$    \\
\begin{tabular}[c]{@{}l@{}}FM* {\scriptsize \citet{liu2022rectified}} \\ {\scriptsize \citet{lipman2022flow}}\end{tabular}& $100$   & $2.96$  & $100$  & $3.05$   \\
\cmidrule{2-5}
\textbf{Energy} \\
\begin{tabular}[c]{@{}l@{}}EDLEBM \\ {\scriptsize \citet{schroder2024energy}}\end{tabular} & $500$  & $73.58$ & $500$  & $36.73$   \\
\begin{tabular}[c]{@{}l@{}}CDLEBM \\ {\scriptsize \citet{pang2020learning}}\end{tabular} & $500$   & $70.15$ & $500$  & $37.87$   \\
\begin{tabular}[c]{@{}l@{}}DRL {\scriptsize \citet{gao2020learning}}\end{tabular}  & $180$   & $9.58$ & $180$  & $5.98$   \\
\begin{tabular}[c]{@{}l@{}}HDEBM {\scriptsize \citet{hill2023tackling}}\end{tabular} & $75$   & $8.06$ & $115$  & $4.13$    \\
E-DSM*  & $35$  & $6.17$  & $79$  & $2.87$   \\
\begin{tabular}[c]{@{}l@{}}CDRL  {\scriptsize \citet{zhu2023learning}}\end{tabular}  & $90$   & $4.31$  & $--$  & $--$   \\
Ours*         & $35$  & $3.01$  & $79$  & $2.60$   \\
\bottomrule
\end{tabular}
\end{center}
\end{table}
\begin{table}[ht]
\caption{Conditional performance by NFE and FID for CIFAR-10 and CelebA.* denotes our result. EDM was used for the teacher score.} \label{tab:conditional}
\begin{center}
\begin{tabular}{lcccc}
\toprule
\multirow{2}{*}{\textbf{Method}} & \multicolumn{2}{c}{\textbf{CIFAR-$10$}} & \multicolumn{2}{c}{\textbf{CelebA-$64$}} \\
 & NFE & FID   & NFE & FID\\
\cmidrule{2-3} \cmidrule{4-5}
\textbf{Diffusion / Flow} \\
\begin{tabular}[c]{@{}l@{}}EDM*  {\scriptsize \citet{karras2022elucidating}}\end{tabular} & $35$   & $1.90$   & $79$  & $1.88$    \\
\begin{tabular}[c]{@{}l@{}}FM* {\scriptsize \citet{liu2022rectified}} \\ {\scriptsize \citet{lipman2022flow}}\end{tabular}& $100$   & $2.69$  & $100$  & $2.95$   \\
\cmidrule{2-5}
\textbf{Energy} \\
E-DSM*  & $35$  & $4.49$  & $79$  & $2.03$   \\
Ours*         & $35$  & $2.71$  & $79$  & $1.95$   \\
\bottomrule
\end{tabular}
\end{center}
\end{table}
\begin{table}[ht]
\caption{Unconditional performance by NFE and FID for AFHQv2 and FFHQ $64$. * denotes our result. EDM was used for the teacher score.} \label{tab:afhq}
\begin{center}
\begin{tabular}{lcccc}
\toprule
\multirow{2}{*}{\textbf{Method}} & \multicolumn{2}{c}{\textbf{AFHQ-64}} & \multicolumn{2}{c}{\textbf{FFHQ-64}} \\
 & NFE & FID  & NFE & FID  \\
\cmidrule{2-3} \cmidrule{4-5}
\textbf{Diffusion / Flow} \\
\begin{tabular}[c]{@{}l@{}}EDM*  {\scriptsize \citet{karras2022elucidating}}\end{tabular} & $79$   & $2.35$  & $79$  & $2.61$     \\
\begin{tabular}[c]{@{}l@{}}FM* {\scriptsize \citet{liu2022rectified}} \\ {\scriptsize \citet{lipman2022flow}}\end{tabular}  & $100$   & $2.73$ & $100$  & $3.30$  \\
\cmidrule{2-5}
\textbf{Energy} \\
E-DSM*  & $79$  & $4.57$  & $79$  & $2.71$ \\
Ours*          & $79$  & $3.88$  & $79$  & $2.64$ \\
\bottomrule
\end{tabular}
\end{center}
\end{table}
\vspace{-0.5cm}
\subsection{Composition of Image Models} \label{exp:img_comp}
As detailed in \Cref{app:experiments}, we train separate energy-parameterized diffusion models for subsets of the CelebA dataset with attributes \textit{male} and \textit{glasses}, $p^{(\text{male})}_{t}$ and $p^{(\text{glasses})}_{t}$. We compose pretrained models via SMC detailed in \Cref{sec:composition_scores}, using potential 
$G_i = \left(\tfrac{p^{(\text{male})}_{t_i}p^{(\text{glasses})}_{t_i}}{M_{t_i}}\right)^{\gamma_{t_i}}$ with schedule $(\gamma_t)_t$ to control diversity.
\begin{figure}[hb]
    \centering
        \includegraphics[trim={3.cm 1.2cm 3cm 1.5cm},clip,width=0.75\linewidth]{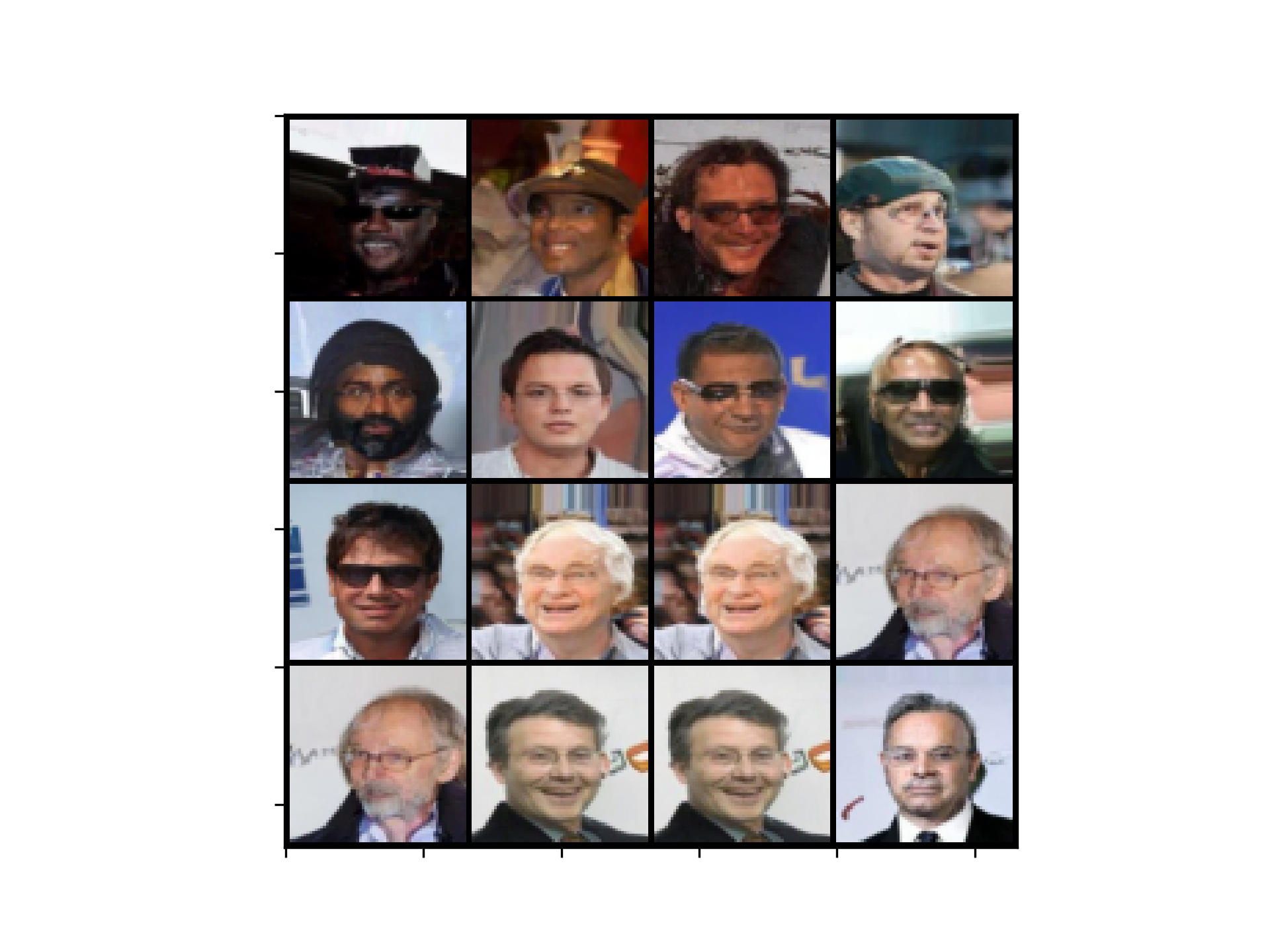}
    \caption{Composition: Male \textsc{AND} Glasses.} 
    \label{fig:celeba_composition}
\end{figure}
\Cref{fig:celeba_composition} shows a uniform subsample from generated batch of size $64$, qualitatively showing our method works for image datasets. Repetition indicates resampling can reduce diversity however.

\subsection{Low Temperature Sampling}\label{exp:low_temp}
We train a conditional energy-diffusion model on CelebA, generate $128$ samples for multiple conditions using a base sampler and via low temperature (temp.) SMC with inverse temp. parameter $\gamma_t=0.1$, then assess images-condition adherence using a CLIP score, detailed in \Cref{app:experiments}. \Cref{tab:clip_celeba} shows the superior performance of low temp. SMC sampling for condition adherence.
\begin{table}[hb]
  \caption{Measuring condition adherence with CLIP.}
  \label{tab:clip_celeba}
  \centering
\begin{tabular}{lcc}
\toprule
\textbf{Condition} & \textbf{Low Temp.} &  \textbf{Base} \\ \hline
{Man with black hair}         & 0.75            & 0.71          \\
{Blonde woman, lipstick}   & 0.63            & 0.53         \\
{Smiling old man}       & 0.95              & 0.85               \\
{Young woman, no hair}  & 0.97       & 0.91   \\
{Man with make-up} & 0.40       & 0.35   \\
\bottomrule
\end{tabular}
\end{table}

\section{Discussion}

\subsection{Related Prior Work}

\textbf{Training Energy Based Models}.
A number of recent works aim to improve EBM training. \citet{zhu2023learning} uses a diffusion model to reduce the number of Langevin steps within the recovery likelihood approach of \cite{gao2020learning}. \cite{schroder2024energy} eliminates the need for MCMC and $\nabla_x$-computation of the energy during training by using a contrastive loss with forward noising process instead of MCMC, coined Energy Divergence (ED). ED is a promising alternative to E-DSM and has connections to score-matching, but, as shown in \Cref{tab:unconditional_gen}, it is not yet competitive, and  suffers from a bias by choice of noisy energy function.

\textbf{Composition with MCMC}. Similar to our work, \cite{du2023reduce} also uses energy-parameterized diffusion models but performs controlled generation with MCMC rather than SMC. SMC is known to suffer weight degeneracy high dimension, resulting in lack of diversity across particles, MCMC does not suffer from this, though requires additional non-parallel steps which is time consuming. The approaches are however complementary, and indeed one may perform MCMC after resampling steps to promote diversity.

\textbf{Sequential Monte Carlo in Diffusion Models.}
Many recent works use SMC within diffusion models for conditional generation, we detail the FKM formulations of these works in in \Cref{app:fkm_potentials}.

\cite{wu2024practical} uses twisted SMC with a classifier-guided proposal \citep{dhariwal2021diffusion} and potentials approximated with diffusion posterior sampling \citep{chungdiffusion}, which has been detailed as a FKM by concurrent work \citep{zhao2024conditional}. \cite{cardoso2024monte} and \cite{dou2024diffusion} tackle linear inverse problems where potentials have a closed form using Gaussian conjugacy. \cite{li2024derivative} perform SMC for both discrete and continuous diffusion models whereby potentials consist of a reward function applied to $\mathbb{E}[\bfX_0|x_t]$.

\cite{liu2024correcting} corrects conditional generations using an adversarially trained density ratio potential, and scales this to text-to-image models.

\textbf{SMC for LLMs}. SMC is not only popular within diffusion models, but has been successful within large language models (LLMs) \citep{lew2023sequential, zhao2024probabilistic}. \cite{lew2023sequential} uses a FKM formulation with indicator based potential functions similar to as detailed in \Cref{sec:bounded}, and \cite{zhao2024probabilistic} discuss using SMC for text using potentials from reward functions or learning such potentials via contrastive twist learning, similar to contrastive learning for EBMs.

\subsection{Concurrent work} Since submission/ acceptance of our work \footnote{Submission October 2024}, there have been a number of relevant concurrent works. 

\textbf{FKM Interpretation}. \cite{singhal2025general} similarly to \cite{zhao2024conditional} and this work, detail sampling diffusion models in terms of KFM. \cite{singhal2025general} follow \cite{li2024derivative} in using reward functions based potentials but focus on text-to-image reward, and explore further heuristics such as or combining rewards via sum or max; and sampling $\bfX_0|x_t$ via nested diffusion \citep{elata2024nested} as input to their reward rather than using $\mathbb{E}[\bfX_0|x_t]$ as done in \cite{li2024derivative}.

\textbf{SMC for discrete diffusion}.  \cite{lee2025debiasing} use SMC for low temperature sampling for discrete diffusion models. \cite{xu2024energy} use a pretrained autoregressive likelihood model applied to samples $\bfX_0|x_t$ for a potential within discrete diffusion sampling.

\textbf{Composition}. \cite{skreta2024superposition} construct a cheap density estimator by simulating from an SDE, which can be computed at sampling time if using reverse diffusion solver, though it is not clear if this can be used in conjunction with resampling and Langevin corrector schemes. \cite{skreta2024superposition} then use this estimator to perform composition-type sampling, however their logical \textsc{AND} appears to differ from other more commonly used logical \textsc{AND} operations, in that it targets samples with equal probability between classes rather than generating both classes, e.g. "a CAT and a DOG" results in a cat/dog hybrid optical illusion rather than a separate cat and separate dog in one image. 

\cite{bradley2025mechanisms} explore composition more formally, establishing types of composition and cases where summing scores is sufficient without need for SMC correction as performed in this work or with MCMC correction from \citep{du2023reduce}.

\subsection{Limitations} 

\textbf{Multiple-networks}. Our distillation loss requires access to pretrained models. Exploring multi-headed networks for joint training of both score and energy were may be an interesting direction to pursue, this would avoid the need for pretrained networks and reduce NFE at sampling time.

\textbf{Diversity}. Resampling may result in a loss of diversity for poorly constructed potentials and proposals. Investigating approximate resampling techniques which preserve diversity such as in \citep{corenflos2021differentiable, ma2020particle,zhu2020towards} may be practical mitigation strategy.

\textbf{Mixing of Scores}. Score-matching and hence our distillation loss may suffer practical issues for supports with isolated components, resulting in learning the incorrect mixing proportions \citep{wenliang2020blindness} - known as the blindness of score-matching. Whilst this may not pose an issue in our setting for $t>0$ due to Gaussian noise connecting the support, there may be issues in using energy functions trained with score matching very close to $t=0$. 

\subsection{Future Considerations} 
\textbf{Scale and modalities.} Whilst we have successfully demonstrated our methods on medium size image datasets, we are yet to verify the performance on other modalities or larger datasets. A first step would be to apply this to latent space \citep{vahdat2021score, rombach2022high} for higher resolution images. Similarly, the energy function is modality agnostic and could be applied to other fields such as molecular dynamics, \citep{arts2023two}.

\textbf{Other application of the energy function}. There are a plethora of applications requiring the energy worth exploring, for example the \textsc{NEGATION} or \textsc{UNION} operations also require access to time-indexed densities \citep{duvisual, koulischer2024dynamic}. Similar to classical EBMs, our learnt energy functions may also be used for unsupervised learning \citep{comet} and reasoning tasks \citep{energy_reasoning}.

\textbf{The benefit of conservative scores}. Although conservative score approximations are not strictly necessary for generative modeling \citep{horvat2024gauge}, our method does enable one to learn SOTA performant diffusion models with strictly conservative scores. Conservative scores have been remarked as crucial in molecular dynamics \citep{arts2023two}; as well as provide attractive theoretical properties \citep{daras2024consistent} in terms of generalization. 

Given the pursuit optimal transport (OT) has attracted a lot of attention \citep{debortoli2021diffusion, thornton2022riemannian, liu20232, shi2023diffusion}, it is worth remarking that strictly conservative drifts are required to recover OT with ReFlow \citep{liu2022rectified}.

\subsubsection*{Acknowledgements}
We thank Rob Brekelmans for feedback on an early draft of this paper and Kevin Li for references \citep{horvat2024gauge, wenliang2020blindness} and fruitful discussion on the advantages/ disadvantages of conservative scores.

\newpage
\bibliographystyle{abbrvnat}
\bibliography{references.bib}



\newpage
\appendix

%
%





%

%

\onecolumn
\runningtitle{Supplementary Material}

\section{On the conservativity of score networks}\label{app:conservative}

\subsection{Conservative Projection}
The conservative projection detailed in \Cref{sec:training_methods} results in the Helmholtz decomposition of the pre-trained score vector field. Generally, consider a vector field $v_t: \mathcal{X} \rightarrow \mathcal{X}$, providing certain conditions hold \citep{liu2022rectified}, then one may decompose $v_t(x_t)=\nabla_x f_t(x_t)+ r_t(x_t)$ where $r_t: \mathcal{X} \rightarrow \mathcal{X}$, $f_t(x_t): \mathcal{X} \rightarrow \rset $ and $\nabla \cdot r_t =0$.

\citet[Theorem 5.2]{liu2022rectified} proves a more general case coined the Bregman Helmholtz decomposition for convex $c: \mathcal{X} \rightarrow \rset$ and conjugate $c^*(x) := \sup_y\{x\cdot y - c(y)\}$, that the optimal $f^*$ of \eqref{eq:convexloss} for vector field $v_t$: 
\begin{align}\label{eq:convexloss}
    \inf_f \int c(v_t(X_t))-v_t(X_t)\cdot g_t(X_t) + c^*(g_t(X_t))\mathrm{d}t && g_t = \nabla c^* \odot \nabla f_t
\end{align}
yields an orthogonal decomposition: $v_t = \nabla c^* \odot \nabla f^*_t + r_t$ where $r_t$ is measure preserving. This implies that given the same initialization stochastic processes defined by vector fields $v_t$ and $\nabla c^* \odot \nabla f^*_t$ have the same marginals.

Our particular case \eqref{eq:conservative_proj} is a specific case of minimizing \eqref{eq:convexloss} for squared Euclidean cost $c(x)=c^*(x)=\tfrac{1}{2}\|x\|^2$; $\nabla c^*(x)=x$. Ideally, if it were available we would use $v_t(x_t)=\nabla \log p_t(x_t)$; but in practice we use $v_t(x_t)=s_\theta(x_t,t)$, for some pre-trained score function $s_\theta$. This specific case of flow matching on a score vector field is also remarked in \citep[Sec 5]{liu2022rectified}, but in a different context.

\subsection{Measuring conservativity}
\begin{figure}[H]
    \centering
    \includegraphics[trim={0 0.1cm 0 0},clip,width=0.3\linewidth]{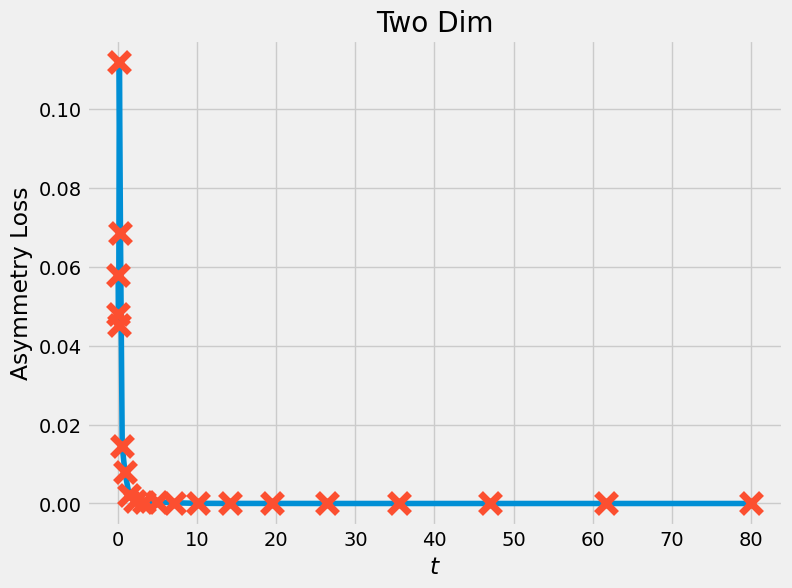}
    \includegraphics[trim={0 0.1cm 0 0},clip,width=0.3\linewidth]{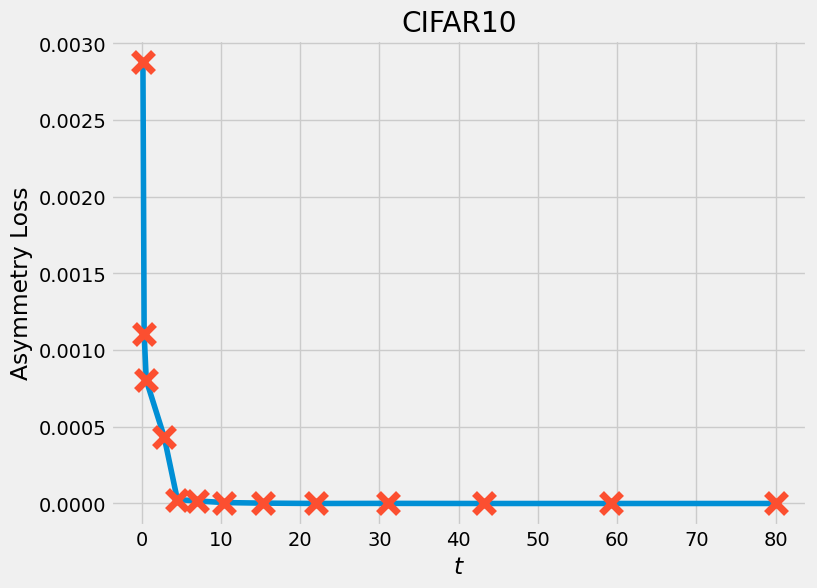}
    \includegraphics[trim={0 0.1cm 0 0},clip,width=0.3\linewidth]{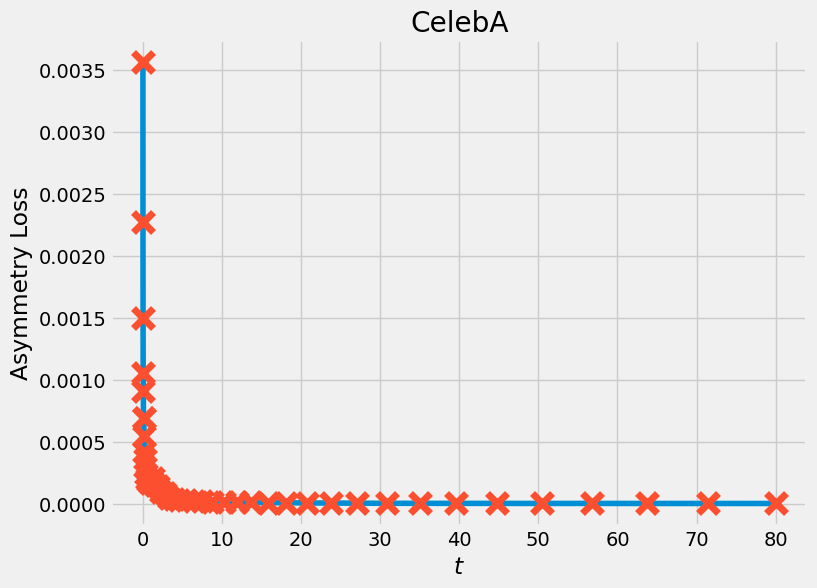}
    \caption{Asymmetry metric  approximation  $\tfrac{\|\jac - \jac^T\|^2}{\|\jac\|^2}\approx \tfrac{\sum_{i=1}^n \|\nu_i^T\jac - \jac\nu_i\|^2}{\sum_{i=1}^n \|\nu_i^T\jac\|^2}$ for $\nu_i \sim \mcn(\mathbf{0}, \Ibb)$ where $\jac=\mathbf{D}_x s_\theta(x_t,t)$ of score network $s_\theta$ trained via DSM on 2D spiral (left) CIFAR1O (middle) and CelebA (right).} 
    \label{fig:conservat}
\end{figure}

By Poincare's star shaped lemma, if a function $g: \rset^d \rightarrow \rset^d $ on star-shaped support has a symmetric Jacobian, then there exists function $F: \rset^d \rightarrow \rset$ such that $g = \nabla F$. Let $\jac$ denote some Jacobian matrix. The Jacobian is typically too large and expensive to compute in full, instead the asymmetry of the Jacobian may be approximated efficiently using Hutchinson's trace estimator: $\|\jac - \jac^T\|^2 = \expect_{\nu \sim \mcn(\mathbf{0}, \Ibb)}\|\nu^T\jac - \jac\nu\|^2$,
\begin{align*}
    \textrm{trace}\left((\jac - \jac^T)^T(\jac - \jac^T)\right) 
    & = \expect_{\nu \sim \mcn(\mathbf{0}, \Ibb)} \nu^T\left(\jac - \jac^T)^T(\jac - \jac^T)\right)\nu \\
    &= \expect_{\nu \sim \mcn(\mathbf{0}, \Ibb)}\|\nu^T(\jac - \jac^T)\|^2 \\
    &= \expect_{\nu \sim \mcn(\mathbf{0}, \Ibb)}\|\nu^T\jac - \nu^T\jac^T\|^2 \\
    &= \expect_{\nu \sim \mcn(\mathbf{0}, \Ibb)}\|\nu^T\jac - \jac\nu\|^2
\end{align*}
where $\nu^T\jac$ and $\jac\nu$ may be computed efficiently with vector-Jacobian and Jacobian-vector products. A Monte Carlo approximation is used in \Cref{fig:afhq_asymmtery} $\tfrac{1}{n}\sum_{i=1}^n \|\nu_i^T\jac - \jac\nu_i\|^2$, and \Cref{fig:conservat} is normalized.

\section{Feynman Kac Model Potentials}\label{app:fkm_potentials}
We detail here a few other Feynman Kac Model (FKM) potentials, alluded to in \Cref{sec:training_methods}.

\subsection{Bounded Generation}
By setting $G_i(x, y) = \mathbb{I}_B(x)$, one forces the generative trajectory to be within region $B\subset \mathcal{X}$.

\begin{figure}[H]
    \centering
    \includegraphics[width=0.4\linewidth]{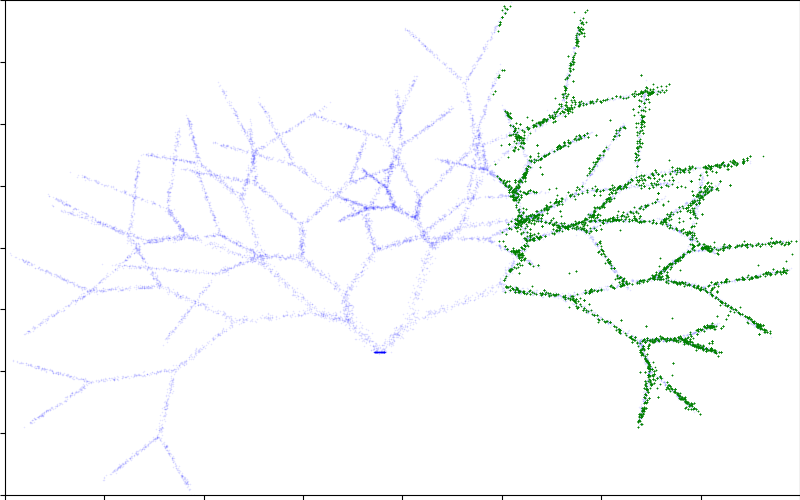}
    \includegraphics[width=0.4\linewidth]{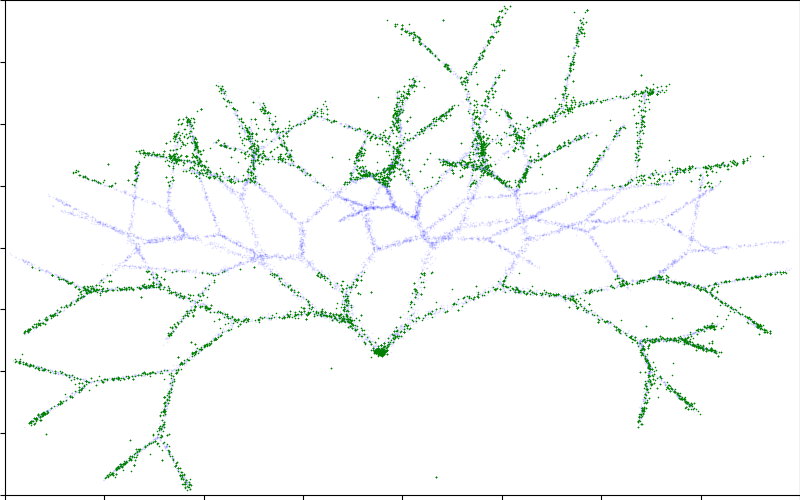}
    \caption{SMC sampling of 2D fractal dataset with FKM potential $G_i(x, y) = \mathbb{I}_B(x)$. Left: $B=[0.25,1] \times \rset$. Right: $B=\rset\times[0.25,1] \cup \times[-1.,-0.1]$. Here blue faded dots are the ground truth data and green points are generated by SMC for the FKM.}
    \label{fig:bounded}
\end{figure}

\subsection{Twisted SMC for Conditional Generation}
\cite{wu2024practical} consider the setting of a pretrained unconditional diffusion model and with to sample from a conditional model. They first use classifier guidance whereby the guidance term is approximated using DPS \citep{chungdiffusion}, then use SMC to correct guided diffusion models.
In the ideal setting the corresponding FKM may be expressed as:
\begin{align}
    M(x_{t_{i+1}}| x_{t_{i}}) &= q^{\lambda}(x_{t_{i+1}}|x_{t_{i}}) \\
    G_0(x_{t_{0}}) &= p(y|x_{t_{0}}) \\
    G_t(x_{t_{i+1}}, x_{t_{i}}) &= \frac{p(y|x_{t_{i+1}})q^\lambda (x_{t_{i+1}}|x_{t_{i}})}{p(y|x_{t_{i}})q^{\lambda}(x_{t_{i+1}}|x_{t_{i}},y)}
\end{align}
where $p(y|x_t)$ denotes a classifier on noisy data for label $y$.

\cite{wu2024practical} assumes the setting where one does not have access to $p(y|x_t)$ or $q^{\lambda}(x_{t_{i+1}}|x_{t_{i}},y)$ but has access to a trained classifier $p_\theta(y|x_0)$ on data without noise; a rained denoiser $D_\theta$, found from the score model, and trained unconditional score model $s_\theta$ to approximate $q^{\lambda}(x_{t_{i+1}}|x_{t_{i}})$ as follows:

\begin{align}
    p^{DPS}_\theta(y|x_{t_{i}}) &= p_\theta(y|D_\theta(x_{t_{i}},t_{i})) \\
    s^{DPS}_\theta(x_{t_{i}},t_{i}, y) &= s_\theta(x_{t_{i}},t_{i})+\nabla_{x_t}\log p^{DPS}_\theta(y|x_{t_{i}}) \\ 
    q_\theta^{DPS,\lambda}(x_{t_{i+1}}|x_{t_{i}},y ) &= \mathcal{N}(x_{t_{i}}+ \Delta_i \frac{\lambda^2+1}{2}[-f(t_{i})x_{t_{i}}+g^2(t_{i})s^{DPS}_\theta(x_{t_{i}},t_{i}, y)], \Delta_i \lambda^2 g(t_{i})^2\mathbb{I})
\end{align}

This yields the guided FKM model:
\begin{align}
    M(x_{t_{i+1}}| x_{t_{i}}) &= q_\theta^{DPS,\lambda}(x_{t_{i+1}}|x_{t_{i}},y) \\
    G_0(x_{t_{0}}) &= p^{DPS}_\theta(y|x_{t_{0}}) \\
    G_t(x_{t_{i+1}}, x_{t_{i}}) &= \frac{p^{DPS}_\theta(y|x_{t_{i+1}})q_\theta^\lambda (x_{t_{i+1}}|x_{t_{i}})}{p^{DPS}_\theta(y|x_{t_{i}})q_\theta^{DPS,\lambda}(x_{t_{i+1}}|x_{t_{i}},y)}.
\end{align}

\section{Sampling Diffusion Models}\label{app:diffusion_samplers}

Let the step size be denoted $\Delta_i = (t_{i+1}-t_{i})$. The simplest implementation of the reverse process \eqref{eq:time_reversed} is:
\begin{align}
    q_\theta^\lambda (x_{t_{i+1}}|x_{t_{i}}) &= \mathcal{N}(x_{t_{i}}+ \Delta_i [-f(t_{i})x_{t_{i}}+g^2(t_{i})\frac{\lambda^2+1}{2}s_\theta(x_{t_{i}},t_{i})], \Delta_i \lambda^2 g(t_{i})^2\mathbb{I})
\end{align}
There exist other more advanced solvers including DDIM (stochastic) \citep{song2020denoising} and a plethora of ODE solvers and higher order SDE and ODE solvers such as the Heun solver used in \citet{karras2022elucidating}.

\section{Composition}\label{app:composition}
Logical compositional operations - \textit{AND}, \textit{OR}, \textit{NOT} - of EBMs may be implemented by transforming the density for EBMs as follows \citep{du2023reduce, duvisual, compose_diffusion}:

\textbf{AND}: $p^\text{AND} \propto \prod_i p^{(i)}$. \textbf{OR}: $p^\text{OR} \propto \sum_i p^{(i)}$. \textbf{NOT}: $p^\text{NOT} \propto \frac{p^{(1)}}{(p^{(2)})^\alpha}$, for some weight $\alpha>0$, depending on density.

Hence one may approximately sample the densities corresponding to each operation via Langevin dynamics. The score is sufficient for \textbf{AND} operations, but the density itself is needed for \textbf{OR} and \textbf{NOT} operations. However, due to score matching learning unnormalised densities, summing such densities for \textbf{OR} composition may be theoretically problematic. In practice this can be resolved with heuristic weighting.

As noted by \citet{du2023reduce}, arithmetic operations of scores at $t>0$ do not in general recover the noisy score corresponding to the composed scores at $t=0$; i.e. $p^\text{AND}_t = \prod_i p^{(i)}_t \neq \int (\prod_i p^{(i)}) \mathrm{d}p_{t|0}$.

 This can lead to a failure in reverse diffusion for compositional generation. One may instead use annealed Langevin dynamics to target $p^\text{AND}_t$ and then gradually anneal $t\rightarrow 0$ to get approximate samples of $p^\text{AND}_0$ \citep{du2023reduce}.

In this work we propose an alternative but complementary approach to annealed Langevin dynamics, in using SMC to target a sequence of distributions which also gradually converge to the desired composition. We specifically target the \textit{AND} operation, but other operations can be targeted similarly. Here the density is required for all logical operations.

Consider schedule $(\gamma_{t_i})_{i=1}^N:$, where $\gamma_{t_N}=1$ and FKM model with potentials:
\begin{align}
    G_i = \tfrac{(p^{(1)}_{t_i}p^{(2)}_{t_i})^{\gamma_{t_i}}M_{t_i}^{1-\gamma_{t_i}}}{M_{t_i}}.
\end{align}
The resulting FKM distributions are
\begin{align}\label{eq:fkm_diff_comp}
    Q(x_{t_{0:N}}) \propto M_{t_{0:N}}(x_{t_{0:N}})G_0(x_{t_0})\prod_{i=1}^{N}G_i(x_{t_i}|x_{t_{i-1}}) 
    = \prod_{i=1}^{N}(p^{(1)}_{t_i}p^{(2)}_{t_i})^{\gamma_{t_i}}M_{t_i}^{1-\gamma_{t_i}} = p^{(1)}_{t_N}p^{(2)}_{t_N} \left[\prod_{i=1}^{N-1}(p^{(1)}_{t_i}p^{(2)}_{t_i})^{\gamma_{t_i}}M_{t_i}^{1-\gamma_{t_i}}\right], 
\end{align}
hence recovers the desired marginal $p^{(1)}_{t_N}p^{(2)}_{t_N}$ at time $t_N$.

\newpage
\section{Experimental Details}\label{app:experiments}
\subsection{Training Details}

\textbf{Compute}. All experiments were carried out using $A100$ 40GB GPUs on single nodes of up to $8$ GPUs. 

\textbf{Training Parameters}
For pre-training via DSM, the learning rates $0.0002$ was used for AFHQv2, CelebA and FFHQ, learning rate $0.001$ is used for CIFAR10. The same learning rates were used for distilling the denoisers into energy-parameterised models.

For E-DSM, $0.0002$ learning rate was used for CIFAR10, with gradient clipping gradients above norm of $10$, other E-DSM clipping was at norm of $1$.

All experiment used a linear warm-up learning rate schedule for $10_000$ steps.

CIFAR10 experiments used batch size $512$, all other image experiments used batch size $256$.

EMA was held constant at a rate of $0.9992$ for all experiments.

\textbf{Network Parameters}:
We used the \textit{SongNet} NCSNPP network as per \citep{karras2022elucidating, song2021scorebased} with the following hyper parameters:
\begin{table}[H]
\caption{Network Parameters.} \label{tab:network_parameters}
\begin{center}
\begin{tabular}{lllll}
\toprule
\textbf{Parameter}                        & \textbf{CIFAR10-32} & \textbf{AFHQV2-64} & \textbf{FFHQ-64} & \textbf{CelebA-64} \\
\textbf{normalization}                    & "GroupNorm"         & "GroupNorm"        & "GroupNorm"      & "GroupNorm"        \\
\textbf{nonlinearity }                   & "swish"             & "swish"            & "swish"          & "swish"            \\
\textbf{nf }                             & 128                 & 128                & 128              & 128                \\
\textbf{ch\_mult }                       & {[} 2, 2, 2{]}      & {[}1, 2, 2, 2{]}   & {[}1, 2, 2, 2{]} & {[}1, 2, 2, 2{]}   \\
\textbf{num\_res\_blocks }               & 4                   & 4                  & 4                & 4                  \\
\textbf{attn\_resolutions  )} & (16,)               & (16,)              & (16,)            & (16,)              \\
\textbf{resamp\_with\_conv  }        & True                & True               & True             & True               \\
\textbf{fir }                            & True                & True               & True             & True               \\
\textbf{fir\_kernel }                    & {[}1, 3, 3, 1{]}    & {[}1, 3, 3, 1{]}   & {[}1, 3, 3, 1{]} & {[}1, 3, 3, 1{]}   \\
\textbf{skip\_rescale }                  & True                & True               & True             & True               \\
\textbf{resblock\_type }                 & "biggan"            & "biggan"           & "biggan"         & "biggan"           \\
\textbf{progressive }                    & "none"              & "none"             & "none"           & "none"             \\
\textbf{progressive\_input }             & "residual"          & "residual"         & "residual"       & "residual"         \\
\textbf{progressive\_combine }           & "sum"               & "sum"              & "sum"            & "sum"              \\
\textbf{attention\_type }                & "ddpm"              & "ddpm"             & "ddpm"           & "ddpm"             \\
\textbf{embedding\_type }                & "fourier"           & "fourier"          & "fourier"        & "fourier"          \\
\textbf{init\_scale }                    & 0.0                 & 0.0                & 0.0              & 0.0                \\
\textbf{fourier\_scale }                 & 16                  & 16                 & 16               & 16                 \\
\textbf{conv\_size }                     & 3                   & 3                  & 3                & 3                  \\
\textbf{num\_scales }                    & 18                  & 18                 & 18               & 18                 \\
\textbf{dropout }                        & 0.13                & 0.25               & 0.05             & 0.1         \\      
\bottomrule
\end{tabular}
\end{center}
\end{table}

\textbf{Sampling}. Generative modelling results using energy parameterized diffusions \cref{tab:conditional}, \cref{tab:unconditional_gen}, \cref{tab:afhq} follow \citep{karras2022elucidating} using the Heun solver on the ODE where $\lambda=0$. 

For low temperature sampling we can also set $\lambda=0$ given the marginals for all $\lambda$ coincide, we do not require to divide by the transition density in the FKM potential. For compositional sampling where we require transition density, we use $\lambda=1$.

\textbf{Evaluation}. Frechet Inception distance \citep{heusel2017gans} was used as the evaluation metric based on Inceptionv3 features \citep{inception} using a re-implementation based on \citep{Seitzer2020FID}.

\newpage
\subsection{Composition}
We train two separate models on subsets of the data, which may overlap for labels "Eyeglasses=1" and "Male=1". We then run SMC using the weighted compositional FKM detailed in \Cref{app:composition}, with $\gamma_t=0.01$; to avoid collapsing to a single sample due to weight degeneracy at time $t=0$; we do not resample for $t<0.1$. It has been observed \citep{karras2024guiding} that conditioning primarily occurs in the middle of the diffusion trajectory and towards $t=0$ all visual features are present but the image is simply sharpened.

The full batch of 64 images, which were then sub-sampled to show in the main document in \Cref{fig:celeba_composition} are given below:
\begin{figure}[h]
    \centering
        \includegraphics[trim={3.cm 1.2cm 3cm 1.5cm},clip,width=0.75\linewidth]{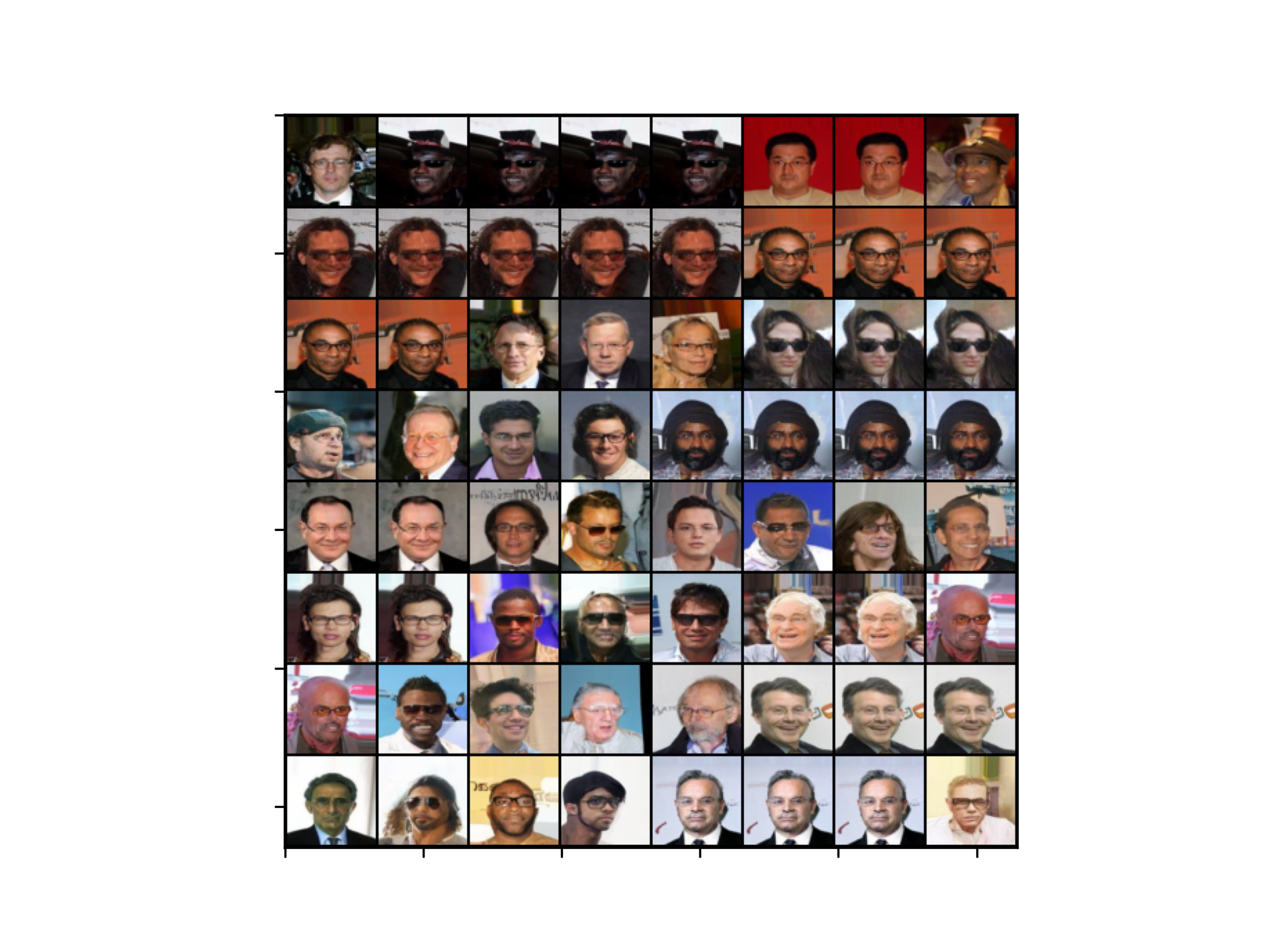}
    \caption{Composition: Male \textsc{AND} Glasses.} 
    \label{fig:celeba_composition64}
\end{figure}

\newpage
\subsection{Low Temperature Generation}
In order to test condition adherence we perform regular sampling of \eqref{eq:time_reversed} with $\lambda=0$ and then using SMC with low temperature weighting setting $\gamma_t=0.1$. We consider the following 5 cases in the \Cref{tab:celeba_attrs}.

\begin{table}[hb]
\caption{CelebA Attribute Conditions} \label{tab:celeba_attrs}
\begin{center}
\small
\begin{tabular}{lccccc}
\toprule
\textbf{Attribute}             & \textbf{Bald Female} & \textbf{Male Make-Up} & \textbf{Smiling Old Man} & \textbf{Blonde Female} & \textbf{Dark Haired Male} \\
\textbf{5\_o\_Clock\_Shadow}   & -1          & -1           & -1              & -1            & -1               \\
\textbf{Arched\_Eyebrows}      & -1          & -1           & -1              & -1            & -1               \\
\textbf{Attractive}            & 1           & 1            & -1              & 1             & 1                \\
\textbf{Bags\_Under\_Eyes}     & -1          & -1           & -1              & -1            & -1               \\
\textbf{Bald}                  & 1           & -1           & -1              & -1            & -1               \\
\textbf{Bangs}                 & -1          & -1           & -1              & -1            & -1               \\
\textbf{Big\_Lips}             & -1          & -1           & -1              & 1             & -1               \\
\textbf{Big\_Nose}             & -1          & -1           & -1              & -1            & -1               \\
\textbf{Black\_Hair}           & -1          & 1            & -1              & -1            & 1                \\
\textbf{Blond\_Hair}           & -1          & -1           & -1              & -1            & -1               \\
\textbf{Blurry}                & -1          & -1           & -1              & -1            & -1               \\
\textbf{Brown\_Hair}           & -1          & -1           & -1              & -1            & -1               \\
\textbf{Bushy\_Eyebrows}       & -1          & -1           & -1              & -1            & -1               \\
\textbf{Chubby}                & -1          & -1           & -1              & -1            & -1               \\
\textbf{Double\_Chin}          & -1          & -1           & -1              & -1            & -1               \\
\textbf{Eyeglasses}            & -1          & -1           & -1              & -1            & -1               \\
\textbf{Goatee}                & -1          & -1           & -1              & -1            & -1               \\
\textbf{Gray\_Hair}            & -1          & -1           & 1               & -1            & -1               \\
\textbf{Heavy\_Makeup}         & -1          & 1            & -1              & 1             & -1               \\
\textbf{High\_Cheekbones}      & 1           & 1            & -1              & 1             & -1               \\
\textbf{Male}                  & -1          & 1            & 1               & -1            & 1                \\
\textbf{Mouth\_Slightly\_Open} & -1          & -1           & -1              & -1            & -1               \\
\textbf{Mustache}              & -1          & -1           & -1              & -1            & -1               \\
\textbf{Narrow\_Eyes}          & -1          & -1           & -1              & -1            & -1               \\
\textbf{No\_Beard}             & -1          & -1           & -1              & -1            & -1               \\
\textbf{Oval\_Face}            & -1          & -1           & -1              & -1            & -1               \\
\textbf{Pale\_Skin}            & -1          & -1           & -1              & -1            & -1               \\
\textbf{Pointy\_Nose}          & -1          & -1           & -1              & -1            & -1               \\
\textbf{Receding\_Hairline}    & -1          & -1           & -1              & -1            & -1               \\
\textbf{Rosy\_Cheeks}          & -1          & -1           & -1              & -1            & -1               \\
\textbf{Sideburns}             & -1          & -1           & -1              & -1            & -1               \\
\textbf{Smiling}               & 1           & 1            & 1               & 1             & 1                \\
\textbf{Straight\_Hair}        & -1          & -1           & -1              & -1            & -1               \\
\textbf{Wavy\_Hair}            & -1          & -1           & -1              & -1            & -1               \\
\textbf{Wearing\_Earrings}     & -1          & -1           & -1              & -1            & -1               \\
\textbf{Wearing\_Hat}          & -1          & -1           & -1              & -1            & -1               \\
\textbf{Wearing\_Lipstick}     & -1          & -1           & -1              & 1             & -1               \\
\textbf{Wearing\_Necklace}     & -1          & -1           & -1              & -1            & -1               \\
\textbf{Wearing\_Necktie}      & -1          & -1           & -1              & -1            & -1               \\
\textbf{Young}                 & 1           & 1            & -1              & 1             & 1         \\      
\bottomrule
\end{tabular}
\end{center}
\end{table}

We then compute the average CLIP score \citep{radford2021learning} for each of a batch of $128$, where the image is up-sampled to $224$, and the following text descriptions are used:
\begin{itemize}
\item "a photo of a young woman with no hair"
\item "a photo of a man wearing make-up"
\item "a photo of an older man who is smiling"
\item "a photo of a blonde woman with lipstick"
\item "a photo of a man with black hair"
\end{itemize}

\subsection{Faster Convergence with Distillation}
Human face datasets are somewhat less diverse than AFHQv2 or CIFAR10, so we notice E-DSM does not struggle as much. Although E-DSM and our distilled approaches yield similar final FID scores for CelebA and FFHQ datasets, the time to train is a significant factor motivating our method. 

Given the careful initialization and pretrained model we notice that our distilled training yields good performance with relatively few training steps. 

With FFHQ-64, after $30,000$ training iterations, E-DSM from scratch yields FID scores of $7.69$, DSM yields FID of $6.18$ and our distilled model yields scores of $3.36$. It takes approx. $200,000$ iterations for E-DSM to reach FID score below $3.3$.

At $30,000$ iterations of training on the unconditional CelebA dataset; E-DSM achieves FID of $5.88$, DSM of $5.18$ and our distilled approach of $2.92$. It takes a further approx. $100,000$ iterations for E-DSM and DSM to reach FID below $3$.

\section{Licences}

\begin{itemize}
    \item JAX Apache-2.0 license \citet{jax2018github}
    \item CelebA: non-commercial research purposes \citep{celeba}
    \item CIFAR-10: MIT license \citep{cifar}
    \item FFHQ: Creative Commons BY-NC-SA 4.0 license \citep{ffhq}
    \item AFHQv2: Creative Commons BY-NC 4.0 license \citep{afhq}
    \item Inception-v3 model: Apache V2.0 license \citep{inception}
    \item CLIP:  MIT license \citep{radford2021learning}
\end{itemize}
\vfill

\end{document}